\documentclass[11pt]{elsarticle}

\usepackage[margin=1in]{geometry}
\usepackage[hidelinks]{hyperref}
\usepackage{amssymb}
\usepackage{enumitem}
\usepackage{booktabs}
\usepackage{caption}
\usepackage{subcaption}
\usepackage{etoolbox}
\usepackage[fleqn]{amsmath}
\usepackage{tikz}
\usepackage{pgfplots}
\pgfplotsset{compat=1.15}
\usetikzlibrary{arrows,automata,pgfplots.fillbetween}
\usetikzlibrary{decorations.pathreplacing,matrix}
\usepgfplotslibrary{statistics}

\usepackage{pifont}

\usepackage{xspace}

\usepackage{siunitx}
\usepackage{mathtools}
\usepackage{stmaryrd}
\usepackage{float}
\usepackage{xcolor}
\usepackage{algorithm}
\usepackage{ragged2e} 
\usepackage{listings}
\usepackage{cleveref}
\usepackage[noend]{algpseudocode}

\usepackage{setspace}
\usepackage{color}

\newtheorem{thm}{Theorem}
\newtheorem{corollary}{Corollary}[thm]

\usepackage{etoolbox}

\makeatletter
\patchcmd{\pprintMaketitle}{\hrule}{\vspace*{-20pt}\hrule}{}{}
\makeatother

\newtheorem{ex}{Example}
\newdefinition{rmk}{Remark}
\newproof{pf}{Proof}
\newproof{pot}{Proof of Theorem \ref{thm2}}
\newcolumntype{L}[1]{>{\raggedright\let\newline\\\arraybackslash\hspace{0pt}}p{#1}}
\newcolumntype{C}[1]{>{\centering\let\newline\\\arraybackslash\hspace{0pt}}p{#1}}
\newcolumntype{R}[1]{>{\raggedleft\let\newline\\\arraybackslash\hspace{0pt}}p{#1}}
\allowdisplaybreaks[1]

\renewcommand{\bfseries}{\fontseries{b}\selectfont}
\robustify\bfseries
\newrobustcmd{\B}{\bfseries}

\newcommand{\blackversion}{ }

\colorlet{myblue}{blue}

\newcommand{\blackorcolor}{
\ifdefined\blackversion
	\color{black}
\fi
}

\newcommand{\REV}[1]{{\color{myblue}\blackorcolor #1}}

\definecolor{red}{rgb}{0.81, 0.06, 0.13}

\definecolor{todored}{rgb}{1, 0, 0}
 	
\biboptions{authoryear}

\journal{arXiv}

\definecolor{psblue}{rgb}{0.36, 0.54, 0.66}
\definecolor{red}{rgb}{0.8, 0.0, 0.0}
\definecolor{green}{rgb}{0.0, 0.5, 0.0}
\definecolor{lila}{rgb}{0.74, 0.2, 0.64}

\begin{document}
	\begin{frontmatter}

		\title{Solving the Paint Shop Problem with Flexible Management of Multi-Lane Buffers Using Reinforcement Learning and Action Masking}
		
		\author[Freiburg]{Mirko Stappert}
		\ead{mirko.stappert@is.uni-freiburg.de}
		\author[Freiburg,Wien]{Bernhard Lutz\corref{cor1}}
		\ead{bernhard.lutz@univie.ac.at}
		\author[VW]{Janis Brammer}
		\ead{janis.adrian.brammer@cariad.technology}
		\author[Freiburg]{Dirk Neumann}
		\ead{dirk.neumann@is.uni-freiburg.de}
		\address[Freiburg]{University of Freiburg, Rempartstr.~16, 79098 Freiburg, Germany}
		\address[Wien]{University of Vienna, Währinger Str.~29, 1090 Vienna, Austria}
		\address[VW]{CARIAD SE, Berliner Ring 2, 38440 Wolfsburg, Germany}
		\cortext[cor1]{Corresponding author.}

\begin{abstract}
In the paint shop problem, an unordered upstream sequence of cars assigned to different colors has to be reshuffled with the objective of minimizing the number of color changes. To reshuffle the upstream sequence, manufacturers can employ a first-in-first-out multi-lane buffer system allowing store and retrieve operations. So far, prior studies primarily focused on simple decision heuristics like greedy or simplified problem variants that do not allow full flexibility when performing store and retrieve operations. In this study, we propose a reinforcement learning approach to minimize color changes for the flexible problem variant, where store and retrieve operations can be performed in an arbitrary order. After proving that greedy retrieval is optimal, we incorporate this finding into the model using action masking. Our evaluation, based on 170 problem instances with 2-8 buffer lanes and 5-15 colors, shows that our approach reduces color changes compared to existing methods by considerable margins depending on the problem size. Furthermore, we demonstrate the robustness of our approach towards different buffer sizes and imbalanced color distributions.
\end{abstract}

\begin{keyword}
Scheduling \sep Paint Shop Problem \sep Reinforcement Learning \sep Metaheuristics  \sep Integer Programming
\end{keyword}
\end{frontmatter}

\onehalfspacing

\vspace{-0.4cm}

\section{Introduction}

The modern automotive industry relies upon a sequential manufacturing process based on assembly lines \citep{Boysen.2022}. One step in this manufacturing process is to color the upstream sequence of cars arriving from the body shop in random order. The sequence of cars is often optimized to minimize work overload or makespan \citep[e.g.,][]{Brammer.2022a,Mosadegh.2020,Neufeld.2023}, typically without accounting for color assignments. However, since color changes of the paint nozzle incur setup costs due to production halts as well as material costs, the paint shop problem focuses on resequencing cars to reduce the number of changes \citep[e.g.,][]{Bysko.2020}. For this purpose, manufacturers can rely on different buffer systems like automated storage and retrieval system \citep[AS/RS, e.g.,][]{Gunay.2017,inman2003algorithm}, pull-off tables \citep[e.g.,][]{Boysen.2011, lahmar2003resequencing}, and multi-lane buffer systems \citep[e.g.,][]{Epping.2003,taube2018resequencing}, as illustrated in \Cref{fig:visualization_form}. The buffer system can be used to store the current car from the upstream sequence to one of the buffer lanes. Each lane operates independently as a first-in-first-out (FIFO) queue.\footnote{In the literature, such buffer systems are also referred to as \emph{selectivity banks}, \emph{parallel line buffer}, or \emph{mix banks}, see \citet{boysen_resequencing_2012}.} In addition, the system can retrieve the car from the rightmost position of a buffer lane, which is then appended to the downstream sequence. The goal of the paint shop problem is to find an optimal sequence of store and retrieve operations that minimizes color changes.


So far, prior studies primarily focused on different problem variants that focus either on the retrieval phase only \citep[e.g.,][]{Epping.2003,hong2018accelerated,spieckermann_sequential_2004} or they require that storage and retrieval are performed in alternating phases, so that the buffer is first fully filled and subsequently fully emptied \citep[e.g.,][]{Ding.2004,wu2021mathematical}. However, it can easily be shown that solving the problem by alternating between filling the buffer first and subsequently removing all stored cars can result in arbitrarily worse solution quality compared to not restricting the order of storage and retrieval. Furthermore, as we show in a later analysis, following the ``store-then-retrieve'' paradigm results in considerably longer sequences of exclusive storage or retrieval operations, thereby preventing a constant flow of cars. We therefore consider the flexible problem variant, which allows store and retrieve operations to be performed at any time point whenever possible.


We propose a deep reinforcement learning (RL) approach for the paint shop problem with flexible multi-lane buffers as they occur at a production facility of our industry partner, a large car manufacturer based in Wolfsburg, Germany. Deep RL appears well suited for scheduling problems due to the discrete action space and time component, allowing us to model the sequential decision-making process \citep[e.g.,][]{Brammer.2022a,Brammer.2022b,Mosadegh.2020}. At each time step, the RL agent can decide between storing the next incoming car in a particular buffer lane or retrieving the rightmost car from a buffer lane. The state representation encodes the buffer content, the next colors from the upstream sequence, and the current color at the reshuffled downstream sequence. The reward function encourages retrieve actions that do not lead to a color change, while color changes are penalized with negative rewards. We consider uniform costs for color changes, in contrast to other studies that account for varying transition costs between specific color pairs \citep[e.g.,][]{Leng.2020,Leng.2023}. This modeling choice is based on the manufacturing process of our industry partner, where setup costs due to production halts significantly outweigh material costs from paint waste. \REV{\citet{Bysko.2020} further note that periodic cleaning of the painting gun is performed regardless of the actual color transition to ensure high coating quality. Manufacturers therefore aim to minimize color changes so that unavoidable transitions can be aligned with these periodic cleaning cycles, thereby reducing avoidable production downtime and costs. In fact, assuming uniform costs of color changes also presents a common choice in the literature \citep[e.g.,][]{Bysko.2020,Huang.2024,taube2018resequencing,wu2021mathematical}.}


We employ action masking to increase the efficiency of the learning progress by excluding invalid actions (i.e., storing in full buffer lanes or retrieving from empty buffer lanes). In addition, we enforce greedy retrieve operations through action masking after showing that greedy retrieval is optimal. Our evaluation based on 170 problem instances with 2--8 buffer lanes and 5--15 colors shows that the proposed RL approach decreases color changes by considerable margins on the vast majority of considered problem instances.
We only observe slightly inferior performance to existing methods on small problems with 2x2 buffers and 15 colors, and
on large problems instances with 8x8 buffers and 10 or 15 colors, which may be explained by the greater complexity of the state space. 
We finally demonstrate the robustness of our approach towards rectangular buffer sizes, initially filled buffer, and problem instances with out-of-distribution characteristics.


\REV{Our study makes two major contributions to the literature. 
First, we formalize the paint shop problem with flexible storage and retrieval operations as an integer linear program (ILP) by generalizing the paint shop problem formulations from \citet{Ding.2004} and \citet{spieckermann_sequential_2004}. To the best of our knowledge, no study has so far provided a complete ILP formalization of the flexible variant. 
We show formally that the solutions of the less flexible problem variant ``storage-then-retrieve'' can be arbitrarily worse regarding the number of color changes than solutions of the flexible problem variant. In addition, we prove optimality of the frequently used greedy retrieval heuristic which aims to retrieve cars of the same color as the first car in the downstream sequence. Second, our study contributes to the growing body of literature applying machine learning to solve problems from operations research \citep{Bengio.2021,Panzer.2022}. The few studies that applied RL to the paint shop problem with multi-lane buffers \citep{Huang.2024,Leng.2020,Leng.2023} restricted flexibility by requiring that store and retrieve operations alternate at each timestep to achieve a desired fill rate (e.g., 60\%). 
Prior studies also proposed several heuristics like greedy storage and greedy retrieval \citep{Bysko.2020,Leng.2020,spieckermann_sequential_2004, sun2015colour}, as well as metaheuristics like genetic algorithms \citep{ko2016paint} and ant colony optimization \citep{Lin.2011}. 
A crucial advantage of RL over metaheuristics and ILP solvers is the fact that a trained RL policy can instantly  generate solutions of decent quality to similar but unseen problem instances. By contrast, metaheuristics and solvers cannot rely on hours of pre-training. Instead, they are built to solve each problem instance individually, independently from other problem instances within the limited cutoff time. RL can transfer prior learning experience to novel instances, which increases its applicability in real-world situations, allowing for fast responses to shot-term disruptions of the production process.}


The remainder of this work is structured as follows. \Cref{sec:rw} provides an overview of related work on the paint shop problem. \Cref{section-PS-problem} formalizes the paint shop problem with fully flexible store and retrieve operations as an integer programming problem. \Cref{section-RL-approach} presents the reinforcement learning approach, describing action space, state representation, transition and reward function, action masking, as well as the policy learning algorithm. \Cref{section-Evaluation} and \Cref{section-Results} detail the evaluation procedure and results. \Cref{section-Conclusion} concludes.

\section{Related work}
\label{sec:rw}


The paint shop problem belongs to the group of resequencing problems.\footnote{The paint shop problem is also known as the color-batching problem \citep[e.g.,][]{Sun.2024, taube2018resequencing}.} Resequencing problems focus on reshuffling a given initial sequence subject to the given flexibility \citep{boysen_resequencing_2012}. Resequencing problems are distinguished between physical resequencing, i.e., work pieces physically change their sequence positions, as well as virtual sequencing, where customer orders are reassigned to work pieces that are physically identical but the upstream sequence does not change physically \citep{Sun.2024}. While virtual resequencing is restricted to matching customer orders to identical work pieces, physical resequencing is restricted by the capabilities of the employed buffer system.

 
\citet{boysen_resequencing_2012} identify three buffer systems for physical resequencing (pull-off tables, AS/RS, and multi-lane buffers) that are frequently studied in the literature. Pull-off tables are placed next to the production line and temporarily remove workpieces from the upstream sequence and reinsert them later at an arbitrary position. AS/RS also offer full flexibility by enabling arbitrary storage and retrieval of workpieces, but they entail significant monetary costs, increased spatial requirements and retrieval time as workpieces have to be lifted up by and put into place by cranes, shuttles, or robotic arms. 
Although pull-off tables and AS/RS both achieve full flexibility, manufacturers generally only implement a handful of pull-off tables compared to hundreds of buffer places of modern AS/RS \citep{boysen_resequencing_2012}. Finally, multi-lane buffers are less flexible than pull-off tables and AS/RS, but they achieve large capacity and fast retrieval times. Apart from the paint shop problem, multi-lane buffers are also employed in car resequencing \citep[e.g.,][]{Boysen.2013,Valero.2014}.


The paint shop problem with multi-lane buffers has been studied in different variants. \Cref{tbl:lit} provides a chronological overview of prior studies, along with the employed solution method, the problem variant restricting the execution of storage and retrieval, and when the solution method changes between storage and retrieval. We identify three distinct problem variants.

First, the ``retrieval-only'' variant considers an initially filled buffer that has to be emptied by retrieve operations. 
\citet{Epping.2003} defined a dynamic programming algorithm inspired by multiple sequence alignment. However, this approach has an exponential time complexity in the number of buffer lanes. Shortly after, \citet{spieckermann_sequential_2004} formulated the retrieval-only problem as an ILP and developed a branch-and-bound approach. The main advantage of this approach is that it can be stopped after a given cutoff time, which increases practicability in the sense that the method finds reasonably good solutions in a feasible time. \REV{The more recent study \citet{hong2018accelerated} revisits dynamic programming approaches and shows that they can be made more efficient by incorporating combinatorial lower bounds and heuristic upper bounds. Their accelerated DP framework thereby reduces the number of states explored and is able to solve instances to optimality for significantly larger problem sizes.}

Second, the ``storage-then-retrieval'' variant alternates between storage and retrieval phases. The solution methods change from storage to retrieval as soon as the buffer is full. The problem variant ``storage-then-retrieval'' was first formalized as an ILP by \citet{Ding.2004}. The authors also suggested greedy heuristics for storage and retrieval. The ILP formulation was later used by \citet{Lin.2011}, who developed a nested ant colony optimization approach. Further heuristics, like the shuffling heuristic and the arraying heuristic procedure, were proposed by \citet{sun2015colour}. The shuffling heuristic is a retrieval heuristic that aims to sequentially merge all buffer lanes by constructing a minimal spanning tree of an associated graph. The array heuristic procedure, by contrast, is a storage heuristic that aims to group together as many cars of the color that occurs most often in the upstream sequence as possible. The recent study by \citet{wu2021mathematical} considers a multi-stage sequencing problem that combines a paint shop with an assembly shop and optimizes the joint problem. 
Some studies \citep[e.g.,][]{Lin.2011,wu2021mathematical} also restrict the length of the upstream sequence to be lower than the buffer size. As a consequence, the solution consists of exactly one storage and one retrieval phase. 

\begin{table}[H]
	\caption{Prior studies on the paint shop problem with multi-lane buffers.\label{tbl:lit}}
	\centering
	\setlength{\tabcolsep}{3pt}
	\footnotesize
	\vspace{-0.2cm}
	\begin{tabular}{L{4.5cm}L{3.5cm}L{4cm}L{3cm}}
		\toprule
		 Study & Solution method  & \multicolumn{2}{c}{Storage and retrieval}  \\
		 \cmidrule(lr){3-4}
		 &  & Variant & Change of phases  \\ \addlinespace
		\midrule
		\citet{Epping.2003}  & Dynamic programming  & Retrieval only / Flexible & {--} / Model decision  \\ \addlinespace
		\citet{spieckermann_sequential_2004} & Branch-and-bound & Retrieval only & {--}  \\ \addlinespace
		\citet{Ding.2004}  & Heuristics &  Store-then-retrieve & Buffer is full	\\ \addlinespace
		\citet{Lin.2011}  & Ant colony optimization & Store-then-retrieve & Buffer is full	 \\ \addlinespace
		\citet{sun2015colour} & Heuristics &  Store-then-retrieve & Buffer is full	 \\ \addlinespace
		\citet{sun2017study}  & Heuristics & Store-then-retrieve & Buffer is full \\   \addlinespace
		\REV{\citet{hong2018accelerated}}  & \REV{Dynamic programming}& \REV{Retrieval only} & \REV{{--}} \\   \addlinespace
		\citet{taube2018resequencing} & ILP-Solver & Multiple buffers & -- \\   \addlinespace
		\citet{Bysko.2020} & Heuristics & Flexible & External signal \\ \addlinespace
		\citet{Leng.2020} & Reinforcement learning & Store-then-retrieve & Each timestep \\ \addlinespace
		\citet{wu2021mathematical} & ILP-Solver & Store-then-retrieve & Buffer is full \\	\addlinespace
		\citet{Leng.2023} & Reinforcement learning & Store-then-retrieve & Each timestep \\ \addlinespace
		\citet{Bysko.2024}  & Game theoretic & Flexible & External signal \\ \addlinespace
		\citet{Huang.2024} & Reinforcement learning & Store-then-retrieve & Each timestep \\ 
		\midrule
		This study  & Reinforcement learning & Flexible & Model decision  \\
		\bottomrule
	\end{tabular}
\end{table}

\vspace{-0.1cm}


Third, the problem variant ``flexible storage and retrieval'' allows store and retrieve operations to be performed in an arbitrary order. The full problem variant was first studied by \cite{Epping.2003}. The authors proposed an exact dynamic programming algorithm for small problem instances with up to 5 colors and 3 buffer lanes. However, the exponential time complexity in the size of the buffer makes this algorithm impractical for any real-sized problem instance. \citet{Bysko.2020} developed heuristic approaches that dynamically assign priorities to certain colors. Although the studies by Bysko and colleagues also considered the flexible problem variant, the decision of whether or not to change from retrieval to storage and vice versa is given by an external signal, e.g., from production and not by the solution approach itself. In their more recent study, \citet{Bysko.2024} developed an approach based on game theory, where the storage and retrieval phases are both formulated as (different) two-player games, where the set of strategies corresponds to the set of buffer lanes. The Nash equilibrium strategy is then proposed as a suitable storage or retrieval operation.


The study by \citet{taube2018resequencing} is conceptually different from other studies and ours. The authors focus on production at the supplier, who generates an internal sequence to optimize three distinct objectives, in particular, minimization of color changes. However, the supplier finally needs to restore the sequence for ``just-in-sequence'' delivery after production as requested by the original equipment manufacturer (OEM). \REV{For this purpose, the supplier employs several multi-lane buffers $m \in \lbrace 2,3,4,5, 10,15 \rbrace$ after assembly line production to ensure that the downstream sequence matches the requested OEM sequence.} The authors emphasize the potential continuity and improvement in the three optimization targets of the approach.


The studies by \citet{Leng.2020}, \citet{Leng.2023}, and \citet{Huang.2024} are closest to ours as they also solved the paint shop problem using deep RL. However, they only train RL for retrieval, while storage is performed by a simple heuristic. In addition, they require that storage and retrieval alternate at each timestep to ensure a given buffer fill rate (e.g., 60\%). By contrast, we consider the fully flexible problem version, where storage and retrieval can be performed in an arbitrary order, while the RL policy learns to perform both, store and retrieve operations.

\section{Paint shop problem}
\label{section-PS-problem}

In this section, we first formalize the paint shop problem. Subsequently, we show that allowing flexibility in performing store and retrieve operations can lead to arbitrarily better solution quality than the less complex problem variant ``store-then-retrieve''.

\subsection{Formalization}

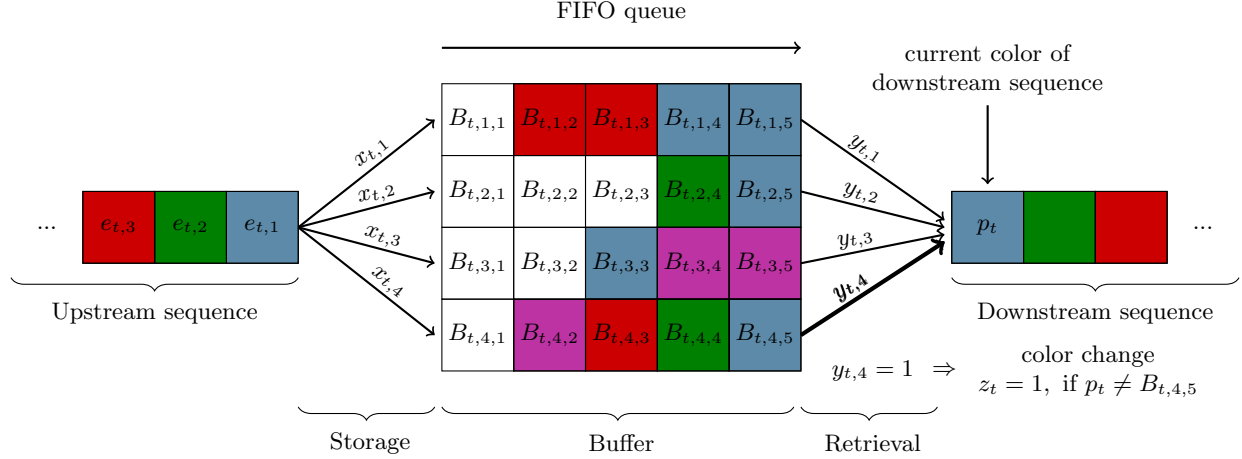
\begin{figure}[H]
\centering
\begin{tikzpicture}[scale=0.95]
\tikzstyle{every node}=[font=\footnotesize]

	\node[text width=3.5cm, align=center] at (10.6,4.2) {current color of\\ downstream sequence};
    \draw[->,thick] (10.6,3.7) -- (10.6,2.6);
    
    \node[] at (-2.5,2) {...};

	
	\draw [fill=psblue] (0,1.5) rectangle ++(1,1) node[midway] {};
    \draw [fill=green](-1,1.5) rectangle ++(1,1) node[midway] {};
    \draw [fill=red](-2,1.5) rectangle ++(1,1) node[midway] {};
    
    \draw  (0,1.5) rectangle ++(1,1) node[midway][scale=1] {$e_{t,1}$};
    \draw (-1,1.5) rectangle ++(1,1) node[midway][scale=1] {$e_{t,2}$};
    \draw (-2,1.5) rectangle ++(1,1) node[midway][scale=1] {$e_{t,3}$};

	\draw [fill=psblue] (10.1,1.5) rectangle ++(1,1) node[midway] {};
    \draw [fill=green](11.1,1.5) rectangle ++(1,1) node[midway] {};
    \draw [fill=red](12.1,1.5) rectangle ++(1,1) node[midway] {};

	\draw[fill=psblue] (7,3) rectangle ++(1,1);   
    \draw[fill=psblue] (6,3) rectangle ++(1,1);  
	\draw[fill=red] (5,3) rectangle ++(1,1);
	\draw[fill=red] (4,3) rectangle ++(1,1); 
	
	\draw[fill=psblue] (7,2) rectangle ++(1,1);  
	\draw[fill=green] (6,2) rectangle ++(1,1); 	
	
	\draw[fill=lila] (7,1) rectangle ++(1,1); 
	\draw[fill=lila] (6,1) rectangle ++(1,1); 	
	\draw[fill=psblue] (5,1) rectangle ++(1,1);
	
	\draw[fill=psblue] (7,0) rectangle ++(1,1); 
	\draw[fill=green] (6,0) rectangle ++(1,1); 	
	\draw[fill=red] (5,0) rectangle ++(1,1);
	\draw[fill=lila] (4,0) rectangle ++(1,1);

    \draw [decorate,
    decoration = {brace, mirror, amplitude=5pt}] (-3,1.3) --  (1,1.3);
    \node[] at (-1,0.8) {Upstream sequence};
 
  \foreach \x in {0,...,4}
    \foreach \y in {0,...,3}
      \pgfmathtruncatemacro\i{4-\y}
      \pgfmathtruncatemacro\j{1+\x}
      \draw (3+\x,\y) rectangle ++(1,1) 
      node[midway][scale=1] {$B_{t,\i,\j}$};

    \draw [decorate,
    decoration = {brace, mirror, amplitude=5pt}] (3,-0.5) --  (7.9,-0.5);
    \node[] at (5.5,-1) {Buffer};
    
	\draw[->,thick]  (3,4.5) -- (8,4.5);
    \node[text width=3cm, align=center] at (5.5,5) {FIFO queue};
    
    \draw [decorate,
    decoration = {brace, mirror, amplitude=5pt}] (1,-0.5) --  (2.9,-0.5);
    \node[] at (2,-1) {Storage};
    
    \draw [decorate,
    decoration = {brace, mirror, amplitude=5pt}] (8,-0.5) --  (9.9,-0.5);
    \node[] at (9,-1) {Retrieval};
    
  \foreach \x in {0,...,3}
  \pgfmathtruncatemacro\j{4-\x}
    \draw[->,thick] (1,2) -- (2.9,\x+0.5) node[pos = 0.6, above, sloped, inner sep=1pt] {$x_{t,\j}$};

  \foreach \x in {0,...,3}
   \pgfmathtruncatemacro\j{4-\x}
    \draw[->, thick] (8,\x+0.5) -- (10,1.8+0.1*\x) node[pos = 0.4, above, sloped, inner sep=1pt] {$y_{t,\j}$};
    \draw[->,ultra thick] (8,0.5) -- (10,1.8) node[pos = 0.4, above, sloped, inner sep=1pt] {$y_{t,4}$};

    \node[] at (9,0) {$y_{t,4}=1$};
    \node[] at (10,0) {$\Rightarrow$};
    \node[align=center] at (12,0) {color change \\$z_t=1, \text{ if } p_t\neq B_{t,4,5}$};
   
\node[] at (13.6,2) {...};
    \draw (10.1,1.5) rectangle ++(1,1) node[midway] {$p_t$};
    \draw (11.1,1.5) rectangle ++(1,1) node[midway] {};
    \draw (12.1,1.5) rectangle ++(1,1) node[midway] {};

    \draw [decorate,
    decoration = {brace, mirror, amplitude=5pt}] (10.1,1.3) --  (14.1,1.3);
    \node[] at (12.1,0.8) {Downstream sequence};
\end{tikzpicture}
\caption{Paint shop problem with a 4x5 buffer ($L=4$ lanes of width $W=5$) and corresponding notation. The binary decision variables are given by $x_{t,i}$ (storage) and $y_{t,i}$ (retrieval).}
\label{fig:visualization_form}
\end{figure}

We formalize the paint shop problem with flexible storage and retrieval operations as an ILP. To ease comprehensibility of the notation, we illustrate an instance of the paint shop problem, along with the problem parameters and variables in \Cref{fig:visualization_form}. An upstream sequence of $N$ cars needs to be painted in their assigned color $c_1, c_2, \dots, c_N$. We introduce a discrete time $t=1,\dots,2N$ to denote the entire sequence of store and retrieve operations. At each timestep $t$, the system can either store the current car from the upstream sequence in one of $L$ buffer lanes or retrieve from a buffer lane. The decision variables $x_{t,i}$ are equal to 1 if the current car is stored in lane $i$ at timestep $t$, and 0 otherwise. A store operation to lane $i$ puts the car in the rightmost empty position of the buffer lane. The color of the car in lane $i$ at position $j$ and timestep $t$ is denoted by $B_{t,i,j}$. The width of each lane is denoted by $W$. The decision variables $y_{t,i}$ are equal to 1 if the last car of buffer lane $i$ is retrieved at timestep $t$. A retrieved car is added to the downstream sequence. The color of the most recently added car is denoted by $p_t$. A color change occurs if the last retrieve operation selected a car at the end of buffer lane $i$ with a different color than $p_t$, which is denoted by the dummy variable $z_t$. Therefore, the goal of the problem is to minimize the sum over all $z_t$ for $t=1,\dots,2N-1$. An overview of all parameters and variables is provided in \Cref{tbl:optim_problem}.

We do not impose length constraints for the downstream sequence. We also do not impose time constraints on the speed of store and retrieve operations. In particular, we assume that painting a car requires more time than storing, moving, and retrieving a car through a buffer lane.

Our formalization extends the formalizations of the problem variants ``store-then-retrieve'' and ``retrieval-only'' studied by \citet{Ding.2004} and \citet{spieckermann_sequential_2004}, respectively.  
The goal \eqref{eqn:target} is to minimize the total number of color changes.
Constraint \eqref{eqn:binary} ensures that the decision variables $x_{t,i},y_{t,i}$ and dummy variables $z_t$ are binary.
Constraint \eqref{eqn:one_operation_at_atime} ensures that exactly one operation is performed at each time step. 
Constraints \eqref{eqn: Sequence_initialization}, \eqref{eqn: Buffer_initialization} and \eqref{eqn: Initialization_currentcolor} initialize the upstream sequence, the buffer, and the current color of the downstream sequence at $t=0$ with zeros. Constraints \eqref{eqn: Sequence_update1} and \eqref{eqn: Sequence_update0} prescribe updates to the upstream sequence: When no store operation is performed, the upstream sequence stays the same, otherwise it is moved by one position. Constraint \eqref{eqn: NoStorage} ensures that once the upstream sequence is empty, no further storing operations are performed. Constraints \eqref{eqn: NoBufferUpdate}-\eqref{eqn: valid_retrieve} describe how store and retrieve operations influence the buffer content. Constraint \eqref{eqn: NoBufferUpdate} ensures that if we neither store in a lane nor retrieve from it, all buffer variables for that lane stay the same. Constraints \eqref{eqn: Buffer_Update_x_rightmost1} and \eqref{eqn: Buffer_Update_x_3} ensure that if we store in a lane, the rightmost free buffer position in that lane is updated to the color of the current car from the upstream sequence. Constraints \eqref{eqn: Buffer_Update_x_rightmost2} and \eqref{eqn: Buffer_Update_x_4} ensure that all other elements of the lane stay unchanged. If we retrieve from a lane, we move all elements one spot to the right \eqref{eqn: Buffer_Update_y}. Constraints \eqref{eqn: valid_store} and \eqref{eqn: valid_retrieve} ensure that when we store in the buffer, the lane is not completely filled and if we retrieve, it is not completely empty. The last three constraints define the current color and color changes. 
Constraint \eqref{eqn: currentcolor_update0} ensures that the current color $p_t$ stays unchanged after a store operation. Constraint \eqref{eqn: currentcolor_update1} ensures that the current color is updated to the color of the corresponding buffer position. Lastly, Constraint \eqref{eqn: Definition_colorchange} states that a color change occurs whenever the colors of two consecutive downstream cars differ.

\begin{table}[H]
\caption{Overview of parameters and variables. \label{tbl:optim_problem}}
\footnotesize
\centering
\vspace{-0.2cm}
\begin{tabular}{L{2.8cm}l}
\toprule
\multicolumn{2}{l}{{\textbf{Parameters}}}\\
\midrule
$C$ & Number of colors\\
$N$ & Length of upstream sequence \\
$c_n \in \{1, \dots, C\}$ & Color of car at initial sequence position $n$\\
$L$ & Number of buffer lanes \\
$W$ & Width of buffer \\ 
\midrule
\multicolumn{2}{l}{{\textbf{Variables}}}\\
\midrule
$x_{t,i}$ & Binary decision variable: 1, if the next incoming car is placed in buffer lane $i$ at time $t$\\
$y_{t,i}$ & Binary decision variable: 1, if the last car in buffer lane $i$ is retrieved at time $t$\\
$B_{t,i,j} \in \{0,\dots,C\} $ & Color of $j$-th car in buffer lane $i$ at time $t$, 0 if empty \\
$e_{t,n}  \in \{0,\dots,C\}  $ & Color of car in position $n$ of upstream sequence at time $t$, 0 if empty\\
$p_t \in \{0,\dots,C\} $ & Current color at time $t$, i.e., color that the last car was painted in \\
$z_t$ & Binary variable indicating if a color change occurs at time $t$\\

\bottomrule
\end{tabular}
\end{table}

\vspace{-0.9cm}
\begin{gather}
\text{Minimize} \sum_{t=1}^{2N-1} z_t \label{eqn:target} \\
\text{Subject to:} \nonumber \\
x_{t,i}, y_{t,i}, z_{t} \in\{0,1\}, \quad t= 1, \dots, 2N; i=1,\dots, L \label{eqn:binary} \\
\sum_{i=1}^L x_{t,i}+y_{t,i} = 1, \quad t=1,\dots, 2N \label{eqn:one_operation_at_atime} \\
%
%
 e_{1,n}=c_n, \quad n=1,\dots, N \label{eqn: Sequence_initialization}\\
 B_{1,i,j}=0, \quad i = 1, \dots, L; j=1, \dots, W \label{eqn: Buffer_initialization} \\
 p_1 = 0 \label{eqn: Initialization_currentcolor} \\
%
%
 \sum_{i=1}^L x_{t,i} =1 \Rightarrow e_{t+1,n}= e_{t,n+1}\wedge e_{t+1,N}=0, \quad t=1, \dots, 2N-1; n=1, \dots, N-1 \label{eqn: Sequence_update1}\\
 \sum_{i=1}^L x_{t,i} =0 \Rightarrow e_{t+1,n}=e_{t,n}, \quad t=1, \dots, 2N-1; n=1, \dots, N \label{eqn: Sequence_update0}\\
 e_{t,1} =0 \Rightarrow x_{t,i}=0, \quad t=1, \dots, 2N; i=1, \dots, L \label{eqn: NoStorage}\\
%
%
 x_{t,i}+y_{t,i}=0 \Rightarrow B_{t+1,i,j} = B_{t,i,j}, \quad t=1,\dots, 2N-1; i=1, \dots, L; j=1,\dots, W  \label{eqn: NoBufferUpdate}\\ 
(x_{t,i}=1 \wedge B_{t,i,W}=0)  \Rightarrow B_{t+1,i,W}=e_{t,1}, \quad t=1,\dots, 2N-1; i=1, \dots, L \label{eqn: Buffer_Update_x_rightmost1}\\
(x_{t,i}=1 \wedge B_{t,i,j}=0 \wedge B_{t,i,j+1}\neq 0 ) \Rightarrow B_{t+1,i,j}=e_{t,1}, \label{eqn: Buffer_Update_x_3}\\
(x_{t,i}=1 \wedge B_{t,i,W} \neq 0)  \Rightarrow B_{t+1,i,W}=B_{t+1,i,W}, \quad t=1,\dots, 2N-1; i=1, \dots, L \label{eqn: Buffer_Update_x_rightmost2}\\
\quad t=1,\dots, 2N-1; i=1, \dots, L; j=1,\dots, W-1 \nonumber \\
(x_{t,i}=1 \wedge B_{t,i,j} \neq 0 \wedge B_{t,i,j+1}=0)  \Rightarrow B_{t+1,i,j}=B_{t,i,j},
\label{eqn: Buffer_Update_x_4}\\
\quad t=1,\dots, 2N-1; i=1, \dots, L; j=1,\dots, W-1  \nonumber\\
y_{t,i}=1 \Rightarrow B_{t+1,i,1}=0 \wedge B_{t+1,i,j+1} = B_{t,i,j}, \quad t=1,\dots, 2N-1; i=1; \dots, L; j=1,\dots, W
\label{eqn: Buffer_Update_y} \\
 x_{t,i}=1 \Rightarrow B_{t,i,1}=0, \quad t=1,\dots, 2N; i=1,\dots, L; j=1,\dots,W  \label{eqn: valid_store}  \\
 y_{t,i}=1 \Rightarrow B_{t,i,W}\neq 0, \quad t=1,\dots, 2N; i=1,\dots, L; j=1,\dots, W  \label{eqn: valid_retrieve}\\
%
%
\sum_{i=1}^L y_{t,i} =0 \Rightarrow p_{t}=p_{t-1}, \quad t=2, \dots, N\label{eqn: currentcolor_update0} \\
y_{t,i} =1 \Rightarrow p_{t}=B_{t,i,W}, \quad t=2, \dots, N; i=1,\dots, L \label{eqn: currentcolor_update1}\\
 |p_{t+1}-p_t|\leq C z_t, \quad t=1, \dots, 2N-1 \label{eqn: Definition_colorchange}
\end{gather}


The implication constraints \eqref{eqn: Sequence_update1}--\eqref{eqn: currentcolor_update1} can be converted into standard linear constraints \citep[see e.g.,][]{Plastria2002formulating}. 
An implication is converted into a linear constraint of the following normal form 
\begin{equation}
a_1=0\wedge\dots\wedge a_n=0\Rightarrow b_1=0\wedge\dots\wedge b_m=0.\label{eqn:normal_form}
\end{equation}
Here, $a_1,\dots,a_n$ are binary variables and $b_1,\dots b_m$ are bounded integer variables with bounds $\mathcal{L}_j \leq b_j \leq \mathcal{U}_j$ for $j=1,\dots,m$. Such a normal form implication constraint is equivalent to the linear inequalities 
\begin{equation}
\mathcal{L}_j\sum_{i=1}^n a_i \leq b_j \leq \mathcal{U}_j\sum_{i=1}^n a_i \text{, for } j=1,\dots,m.
\end{equation}
Each of the constraints \eqref{eqn: Sequence_update1}--\eqref{eqn: currentcolor_update1} can be converted into the normal form \eqref{eqn:normal_form} by introducing further dummy variables. For instance, for constraint \eqref{eqn: Sequence_update1} we define $b_2 = e_{t+1,N}$, $a_1=1-\sum_{i=1}^L x_{t,i}$ and $\ b_1=e_{t+1,n}-e_{t,n+1}$.
The bounds are $\mathcal{L}_1=-C, \mathcal{L}_2=0,\mathcal{U}_1=\mathcal{U}_2=C$. The implication constraint \eqref{eqn: Sequence_update1} is thus equivalent to the following two linear constraints
\begin{align}
-C(1-\sum_{i=1}^L x_{t,i}) & \leq e_{t+1,n}-e_{t,n+1} \leq C(1-\sum_{i=1}^Lx_{t,i}) ,\quad t=1, \dots, 2N-1; n=1, \dots, N-1 \\
0& \leq e_{t+1,N} \leq C(1-\sum_{i=1}^Lx_{t,i}) , \quad t=1, \dots, 2N-1.
\end{align}
Note that Gurobi supports implication constraints, which are called ``indicator constraints'' in the Gurobi manual \citep[p.~515]{gurobi}.

\subsection{Solution quality}
\label{section:Solution-quality}

We now demonstrate that the solution quality of the variant ``store-then-retrieve'' can be arbitrarily worse compared to ``flexible storage and retrieval''. We first provide an example to illustrate the point and subsequently formulate the theorem.

\begin{ex}
\normalfont
Consider a simple problem instance with two colors (coded as 1 and 2), a 2x2 buffer, and the upstream sequence $[2,1,1,1,2,1,1,1]$. We assume that the sequence is processed in the two equal blocks $[2,1,1,1]$ and $[2,1,1,1]$. As a consequence, two iterations of ``store-then-retrieve'' will cause two color changes, one for each block since the last car of color $2$ must also be stored in the buffer. 
However, if we instead generate a sequence with fully flexible store and retrieve operations, we can store and retrieve the first three $1$s without any color change. Subsequently, we store the 2 in the first lane. We then store and retrieve the next three 1s through the second lane. Finally, we store the last 2 and retrieve it with the other 2, which results in only one color change.
\end{ex}
Although this example seems small, it is actually generic in the sense that we can extend the problem instance to obtain an arbitrarily higher number of color changes.
Also note that ``flexible storage and retrieval'' always results in solutions which are as least as good as solutions generated by ``store-then-retrieve''.
\begin{thm}
Consider the paint shop problem with more than one buffer lane $L > 1$, more than one color $C>1$, and lane width $W > 1$. 
For every number $n\in \mathbb{N}$ there is an upstream sequence for which ``store-then-retrieve'' causes at least $n$ more color changes than ``flexible storage and retrieval.''
\end{thm}
\begin{pf}
See Appendix B of the supplementary material.
\end{pf}

\section{Reinforcement learning approach}
\label{section-RL-approach}

We now describe our RL approach. Specifically, we first formulate the paint shop problem as an RL problem, consisting of an environment, state and action space as well as transition and reward function. Subsequently, we describe how action masking is used to incorporate prior human knowledge of optimal strategies. Finally, we explain the employed method (proximal policy optimization) for policy learning.

\subsection{Environment}

The environment models the whole paint shop, including store and retrieve operations which transform the upstream sequence into the downstream sequence. The discrete time $t=1,\dots,2n$ refers to the full sequence of store and retrieve operations. Each learning episode starts with an upstream sequence of length $N$ and an empty buffer. At each time step, the environment provides the RL agent with the current state, which contains the buffer content, the most recent colors from the upstream sequence, and the current color of the last element in the downstream sequence. The RL agent can then either store the current car from the upstream sequence in the rightmost empty position of a buffer lane or retrieve the car from the last position of a buffer lane, which is then appended to the downstream sequence. The environment simulates the performed action by updating the upstream sequence, buffer content, and downstream sequence. A learning episode ends when the initial sequence and buffer are both empty. The environment also calculates the reward for the agent. We implement the environment in Python 3.8.0, using the ``OpenAI Gym'' package in version 0.21.0 \citep{brockman_openai_2016}.

\subsection{State representation}

The state contains the buffer content, the $K$ next colors from the upstream sequence, and the current color of the last car in the downstream sequence
\begin{equation} \label{integer-state-representation}
s_t= (\underbrace{B_{t,1,1},\dots,B_{t,L,W}}_{\text{Buffer content}}, \underbrace{e_{t,1},\dots,e_{t,K}}_{\substack{ \text{K next incoming cars} \\ \text{of input sequence}}}, \underbrace{p_t}_{\substack{ \text{Current} \\ \text{painting color}}}).
\end{equation}
Each value ($0,1,\dots, C$) either denotes an empty position (encoded as 0) or a color (encoded as $1,\dots,C$). We determine the value of the look-ahead length ($K=5$) through a pre-study, see Appendix C of the supplementary material. However, encoding a color by an integer number $\{0,\dots C\}$ suggests that colors can be ordered in an ordinal way from low to high, although such an ordering is not present as colors reflect categorical values. Accordingly, we represent colors using a one-hot encoding, so that each color $c \in \{0,\dots C\}$ is mapped to a vector $o(c)=(0,\dots, 1, \dots,0)$ of length $C$ with a one at position $c$ and zeros elsewhere. The vector consisting of zeros only represents an empty buffer or sequence position. The one-hot state representation is thus given as 
\begin{equation}
s_t^{\text{one-hot}}= (\underbrace{o(B_{t,1}),\dots,o(B_{t,L,W})}_{\text{Buffer content}}, \underbrace{o(e_{t,1}),\dots,o(e_{t,K})}_{\substack{ \text{K next incoming cars} \\ \text{of input sequence}}}, \underbrace{o(p_t)}_{\substack{ \text{Current} \\ \text{painting color}}}). \label{one-hot-state-representation}
\end{equation}


\subsection{Action space and transition function}

At each timestep, the RL agent can either store the next car from the upstream sequence in a buffer lane or retrieve the rightmost car from a buffer lane, appending it to the downstream sequence. Thus, the action space $A$ consists of retrieve ($1,\dots,L$) and store actions ($L+1, \dots, 2L$)
\begin{equation} \label{state-space}
A = \{\underbrace{1,\dots, L}_{\substack{\text{Retrieve}\\ \text{actions}}}, \underbrace{L+1,\dots,2L}_{\substack{\text{Store}\\ \text{actions}}}  \}.
\end{equation}
A retrieve action $a \in \lbrace 1,\dots, L \rbrace$ retrieves the rightmost car from lane $a$, while a store action $a \in \lbrace L+1,\dots, 2L \rbrace$ stores the current car from the upstream sequence in lane $a-L$.

\subsubsection*{Transition function}

The transition function denotes how the paint shop evolves from state $s_t$ to $s_{t+1}$ if a store or retrieve action is performed. If an action is invalid, that is, attempting to store a car in a full lane, store once the upstream sequence is empty, or retrieving from an empty lane, the transition function returns the current state $s_t$.

\subsubsection*{Storage}

A store action $a\in \{L+1,\dots, 2L\}$ stores the current car from the upstream sequence in the rightmost empty position in buffer lane $a-L$ . All cars from the upstream sequence then move one position forward, while the color of the last car in the downstream sequence does not change. The transition function for storage is formalized in \Cref{alg:transition_storage}.

\subsubsection*{Retrieval}

A retrieve action $a\in \{1,\dots, L\}$ appends the rightmost car in lane $a$ to the downstream sequence. The current color is updated to the color of the retrieved car and all other cars of lane $a$ move one position to the right. The upstream sequence does not change. This is formalized in \Cref{alg:transition_retrieval}.

\begin{center}
\begin{minipage}{0.6\textwidth}
\begin{algorithm}[H]
\caption{Transition function: Storage \label{alg:transition_storage}}
\footnotesize
\begin{algorithmic}[1]
\State \textbf{input:} state $s_t$, action $a\in \{L+1,\dots, 2L\}$ // store car in lane $a-L$
\If{$B_{t,a-L,1} \neq 0$ \textbf{ or } $e_{t,1}=0$}
\State \textbf{output:} $s_t$ // invalid action, return current state
\EndIf
\State $p_{t+1} = p_t$ // current color in downstream sequence does not change
\ForAll{$n=2, \dots, N$}
\State$e_{t+1,n}=e_{t,n}$ // update upstream sequence
\EndFor
\State $e_{t+1,N}=0$  // first position becomes empty
\State $j^* = \max_j \text{ so that } B_{t,a-L,j} = 0$
\State $B_{t+1,a-L,j^*} = e_{t,1}$ // car is placed in rightmost available space in lane $a-L$
\ForAll{$j = 1,\dots,W \text{ with } j \neq j^*$}
\State $\ B_{t+1,a-L,j}=B_{t,a-L,j}$ // no changes to other cars in lane $a-L$
\EndFor
\ForAll{$i=1,\dots,L \text{ with } i \neq a-L$} 
\State $\ B_{t+1,i,j}=B_{t,i,j}$ // no changes in other lanes
\EndFor
\State \textbf{output:} $s_{t+1}= (B_{t+1,1,1},\dots,B_{t+1,L,W}, e_{t+1,1},\dots,e_{t+1,K}, p_{t+1})$
\end{algorithmic}
\end{algorithm}
\end{minipage}
\end{center}

\begin{center}
\begin{minipage}{0.6\textwidth}
\begin{algorithm}[H]
\caption{Transition function: Retrieval \label{alg:transition_retrieval}}
\footnotesize
\begin{algorithmic}[1]
\State \textbf{input} state $s_t$, action $a\in \{1,\dots, L\}$ // retrieve from lane $a$
\If{$B_{t,a,W} = 0$}
\State \textbf{output:} $s_t$ // invalid action, return current state
\EndIf
\State$p_{t+1} = B_{t,a,W}$ // current color is updated
\ForAll{$n=1,\dots N$}
\State $e_{t+1,k}=e_{t,k}$  // no change in upstream sequence
\EndFor
\State$B_{t+1,a,1}=0 $      // left-most entry of lane $a$ becomes empty

\ForAll{$j=2, \dots, W$}
\State$B_{t+1,a,j}=B_{t,a,j-1}$ // move cars to the right in lane a
\EndFor

\ForAll{$i\neq a \text{ and } j=1, \dots, W$} 
\State $\ B_{t+1,i,j}=B_{t,i,j}$ // no changes in other lanes
\EndFor
\State \textbf{output:} $s_{t+1}= (B_{t+1,1,1},\dots,B_{t+1,L,W}, e_{t+1,1},\dots,e_{t+1,K}, p_{t+1})$
\end{algorithmic}
\end{algorithm}
\end{minipage}
\end{center}


\subsection{Reward function}

The reward function should guide the RL agent towards performing valid actions, while minimizing the number of color changes. Accordingly, the reward function penalizes invalid actions and retrieve actions causing color changes. Invalid actions are penalized with a reward of $-10$. Valid store operations and retrieve actions causing a color change are assigned a zero reward. However, valid retrieve operations that do not cause a color change yield a positive reward of one. Taken together, the reward function is given as
\begin{equation} 
r(s_{t},a)=\begin{cases}
0, &\text{if } a \leq L \text{ and } B_{t,a,W}\neq 0 \text{ and } B_{t,a,W} \neq p_t \text{ (retrieval with color change)}  \\
1, &\text{if } a \leq L \text{ and } B_{t,a,W}\neq 0\text{ and } B_{t,a,W}=p_t \text{ (retrieval without color change)}\\
-10, &\text{if } a \leq L \text{ and } B_{t,a,W}=0 \text{ (invalid retrieval action)} \\
0, &\text{if } a > L \text{ and } B_{t,a-L,1}= 0 \text{ and } e_{t,1} \neq 0 \text{ (valid store action)}\\
-10, &\text{if } a > L \text{ and } B_{t,a-L,1} \neq 0 \text{ or } e_{t,1} = 0 \text{ (invalid store action)}. 
\end{cases}
\label{reward-function} 
\end{equation}

Note that varying costs of color changes, specified by a cost matrix $CC \in \mathbb{R}_{\geq 0}^{C \times C}$ \citep[e.g.,][]{Leng.2020,Leng.2023}, could easily be incorporated by replacing the reward for ``retrieval with color change'' with the negative value of the corresponding entry $CC[p_t,B_{t,a,W}]$.

\setbool{@fleqn}{true}
\subsection{Properties and action masking}
\label{section:Action-masking}

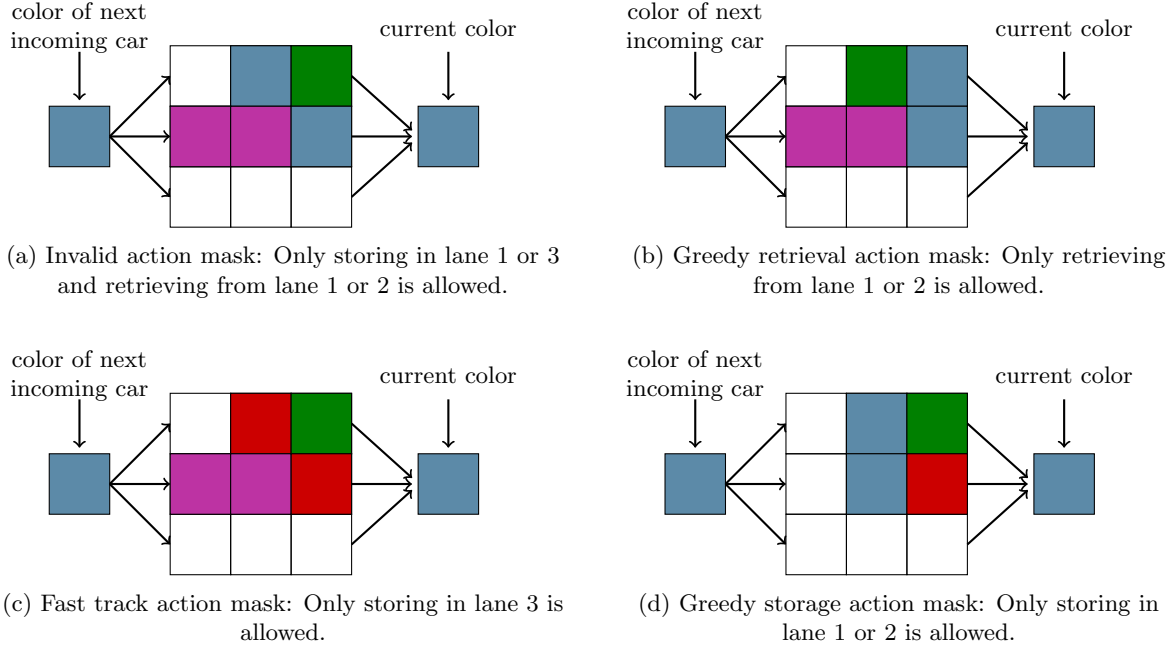
\begin{figure}[H]
\captionsetup[subfigure]{justification=centering,font=footnotesize,skip=0pt}
	\footnotesize
     \centering
     \begin{subfigure}[b]{0.45\textwidth}
         \begin{tikzpicture}[scale=0.8]
         
\tikzstyle{every node}=[font=\footnotesize]
    \node[align=center] at (0.5,3.3) {color of next \\ incoming car};
    \draw[->,thick] (0.5,2.9) -- (0.5,2.1);

    \node[] at (6.6,3.3) {current color};
    \draw[->,thick] (6.6,2.9) -- (6.6,2.1);
    \draw [fill=psblue] (0,1) rectangle ++(1,1) node[midway] {};
\draw [fill=psblue] (6.1,1) rectangle ++(1,1) node[midway] {};  
  \foreach \x in {0,...,2}
    \foreach \y in {0,...,2}
      \pgfmathtruncatemacro\i{3-\y}
      \pgfmathtruncatemacro\j{1+\x}
      \draw (2+\x,\y) rectangle ++(1,1) 
      node[midway] {};
	\draw[fill=psblue] (3,2) rectangle ++(1,1);  
	\draw[fill=green] (4,2) rectangle ++(1,1);
	\draw[fill=lila] (2,1) rectangle ++(1,1); 
	\draw[fill=lila] (3,1) rectangle ++(1,1); 	
	\draw[fill=psblue] (4,1) rectangle ++(1,1);      
   \foreach \x in {0,...,2}
    \draw[->,thick] (1,1.5) -- (2,\x+0.5);  
  \foreach \x in {0,...,2}
    \draw[->,thick] (5,\x+0.5) -- (6,1.4+0.1*\x); 
\end{tikzpicture}
\vspace{0.2cm}
         \caption{Invalid action mask: Only storing in lane 1 or 3 and retrieving from lane 1 or 2 is allowed.}
	     \end{subfigure}
	     \hspace{0.5cm}
     \begin{subfigure}[b]{0.45\textwidth}
         \begin{tikzpicture}[scale=0.8]
\tikzstyle{every node}=[font=\footnotesize]
    \node[align=center] at (0.5,3.3) {color of next \\ incoming car};
    \draw[->,thick] (0.5,2.9) -- (0.5,2.1);
    \node[] at (6.6,3.3) {current color};
    \draw[->,thick] (6.6,2.9) -- (6.6,2.1);
    \draw [fill=psblue] (0,1) rectangle ++(1,1) node[midway] {};
\draw [fill=psblue] (6.1,1) rectangle ++(1,1) node[midway] {};  
  \foreach \x in {0,...,2}
    \foreach \y in {0,...,2}
      \pgfmathtruncatemacro\i{3-\y}
      \pgfmathtruncatemacro\j{1+\x}
      \draw (2+\x,\y) rectangle ++(1,1) 
      node[midway] {};
	\draw[fill=green] (3,2) rectangle ++(1,1);  
	\draw[fill=psblue] (4,2) rectangle ++(1,1);
	
	\draw[fill=lila] (2,1) rectangle ++(1,1); 
	\draw[fill=lila] (3,1) rectangle ++(1,1); 	
	\draw[fill=psblue] (4,1) rectangle ++(1,1);      
   \foreach \x in {0,...,2}
    \draw[->,thick] (1,1.5) -- (2,\x+0.5);  
  \foreach \x in {0,...,2}
    \draw[->,thick] (5,\x+0.5) -- (6,1.4+0.1*\x); 
\end{tikzpicture}
\vspace{0.2cm}
         \caption{Greedy retrieval action mask: Only retrieving from lane 1 or 2 is allowed.}
     \end{subfigure} \\
     
     \vspace{0.5cm}

     \begin{subfigure}[b]{0.45\textwidth}        
        \begin{tikzpicture}[scale=0.8]
\tikzstyle{every node}=[font=\footnotesize]
    \node[align=center] at (0.5,3.3) {color of next \\ incoming car};
    \draw[->,thick] (0.5,2.9) -- (0.5,2.1);
    \node[] at (6.6,3.3) {current color};
    \draw[->,thick] (6.6,2.9) -- (6.6,2.1);
    \draw [fill=psblue] (0,1) rectangle ++(1,1) node[midway] {};
\draw [fill=psblue] (6.1,1) rectangle ++(1,1) node[midway] {};  
  \foreach \x in {0,...,2}
    \foreach \y in {0,...,2}
      \pgfmathtruncatemacro\i{3-\y}
      \pgfmathtruncatemacro\j{1+\x}
      \draw (2+\x,\y) rectangle ++(1,1) 
      node[midway] {};
	\draw[fill=red] (3,2) rectangle ++(1,1);  
	\draw[fill=green] (4,2) rectangle ++(1,1);
	\draw[fill=lila] (2,1) rectangle ++(1,1); 
	\draw[fill=lila] (3,1) rectangle ++(1,1); 	
	\draw[fill=red] (4,1) rectangle ++(1,1);      
   \foreach \x in {0,...,2}
    \draw[->,thick] (1,1.5) -- (2,\x+0.5);  
  \foreach \x in {0,...,2}
    \draw[->,thick] (5,\x+0.5) -- (6,1.4+0.1*\x); 
\end{tikzpicture}
\vspace{0.2cm}
         \caption{Fast track action mask: Only storing in lane 3 is allowed.}
     \end{subfigure}
\hspace{0.5cm}     
     \begin{subfigure}[b]{0.45\textwidth}
         \begin{tikzpicture}[scale=0.8]
\tikzstyle{every node}=[font=\footnotesize]
    \node[align=center] at (0.5,3.3) {color of next \\ incoming car};
    \draw[->,thick] (0.5,2.9) -- (0.5,2.1);
    \node[] at (6.6,3.3) {current color};
    \draw[->,thick] (6.6,2.9) -- (6.6,2.1);
    \draw [fill=psblue] (0,1) rectangle ++(1,1) node[midway] {};
\draw [fill=psblue] (6.1,1) rectangle ++(1,1) node[midway] {};  
  \foreach \x in {0,...,2}
    \foreach \y in {0,...,2}
      \pgfmathtruncatemacro\i{3-\y}
      \pgfmathtruncatemacro\j{1+\x}
      \draw (2+\x,\y) rectangle ++(1,1) 
      node[midway] {};
	\draw[fill=psblue] (3,2) rectangle ++(1,1);  
	\draw[fill=green] (4,2) rectangle ++(1,1);
	
	\draw[fill=psblue] (3,1) rectangle ++(1,1); 
	\draw[fill=red] (4,1) rectangle ++(1,1);      
   \foreach \x in {0,...,2}
    \draw[->,thick] (1,1.5) -- (2,\x+0.5);  
  \foreach \x in {0,...,2}
    \draw[->,thick] (5,\x+0.5) -- (6,1.4+0.1*\x); 
\end{tikzpicture}
\vspace{0.2cm}
     	\caption{Greedy storage action mask: Only storing in lane 1 or 2 is allowed.}
     \end{subfigure}  
        \caption{Illustration of the four considered action masks.}
        \label{fig:Action_mask_illustration}
\end{figure}

The paint shop problem has several useful properties, which we can exploit using action masking. 
An action mask is a function $m{:}\,(s_t,a)\rightarrow \{0,1\}$ that can be employed to reduce the set of admissible actions in a given state. An action $a$ is admissible in state $s_t$ if $m(s_t,a)=1$. Specifically, it may be desired to reduce the set of admissible actions to (i) valid actions, (ii) provably optimal actions, or (iii) actions suggested by heuristics. We thus consider different action masks that enforce valid actions, as well as the provably optimal strategies ``greedy retrieval'' and so-called ``fast-track actions.'' Furthermore, we consider an action mask that enforces greedy store actions. Thereby, we aim to increase the efficiency of the learning process by avoiding invalid actions and directly incorporating prior human knowledge about the problem. An illustration of all considered action masks is provided in \Cref{fig:Action_mask_illustration}. Subsequently, we consider each action mask in detail.

\subsubsection*{Invalid actions}

Our first action mask $m^{\text{INV}}$ excludes invalid actions 
\begin{equation}
m^{\text{INV}}(s_t,a)=\begin{cases}
1\text{, if } a\in\{1,\dots,L\} \text{ and } B_{t,a,W}\neq 0 \\ 
1\text{, if } a\in\{L+1,\dots,2L\} \text{ and }  B_{t,a-L,1}=0 \text{ and } e_{t,1}\neq 0 \\
0, \text{ else}.
\end{cases}
\end{equation} 
Adding this action mask hence waives the need to penalize invalid actions with negative rewards. Thus, the learning episodes can also be expected to be shorter, resulting in a more efficient learning process overall.

\subsubsection*{Greedy retrieval}

\setbool{@fleqn}{false}

An action is called a ``greedy retrieve action'' if it retrieves a car from the buffer that has the same color as the current color. To show that greedy retrieval is optimal, we first introduce the cumulative reward that arises when performing a sequence of actions $a_1,\dots, a_n$ starting from a given state $s_0$ 
\begin{equation}
r(s_0,a_1,\dots, a_n) = \sum_{t=0}^{n-1}r(s_t,a_{t+1}).\label{eqn:cum_reward}
\end{equation}

We call an action $a^*$ \emph{optimal} in state $s_t$ if a sequence that maximizes the cumulative reward in $s_t$ starts with $a^*$, formally 
\begin{equation}
\max_{a_1,\dots, a_n\in A}r(s_t,a^*,a_1,\dots, a_n )\geq  r(s_t,a^\prime,a^\prime_1,\dots, a^\prime_n), \ \forall a^\prime, a^\prime_1,\dots, a^\prime_n \in A.
\end{equation}
Note that there can be multiple optimal actions in a given state. We define the optimality of a sequence of actions accordingly.

\begin{thm}
\label{thm:greedy}
Greedy retrieve actions are optimal.
\end{thm}
\begin{pf}
See Appendix B of the supplementary material.
\end{pf}
Following \Cref{thm:greedy}, we consider the greedy retrieval action mask $m^{\text{GR}}(s_t,a)$
\begin{equation}
m^{\text{GR}}(s_t,a)=\begin{cases}
1\text{, if } a\in\{1,\dots,L\} \text{ and } p_t=B_{t,a,W} \\
1, \text{ if } p_t\neq B_{t,i,W}\ \forall i=1,\dots, L \\
0\text{, else}.
\end{cases}
\end{equation} 
The mask enforces greedy retrieval if it is possible. If greedy retrieval is not possible, the mask allows all actions.

\subsubsection*{Fast-track actions}

We refer to a ``fast-track action'' as storing a car to an empty buffer lane in order to directly retrieve in the next step without causing a color change. Given optimality of greedy retrieval and the fact that store actions do not receive a reward, we can directly derive that fast-track actions are also optimal.
\begin{corollary}
\label{thm:fast-track}
Fast-track actions are optimal.
\end{corollary}
The fast-track action mask $m^{\text{FT}}$ enforces store actions to empty lanes if a subsequent greedy retrieval from this lane is possible, consistent with \Cref{thm:fast-track}. Conversely, the mask allows all actions if a fast-track action is not possible
\begin{equation}
m^{\text{FT}}(s_t,a)=\begin{cases}
1\text{, if } a\in\{L+1,\dots,2L\} \text{ and } B_{t,a-L,1}=0 \text{ and } p_t=e_{t,1}\\
1, \text{ if } p_t\neq e_{t,1} \text{ or } B_{t,i,1}\neq 0, \ \forall i=1,\dots, L \\
0\text{, else}.
\end{cases}
\end{equation}

\subsubsection*{Greedy storage}

Finally, we consider ``greedy storage actions,'' i.e., storing the current car from the upstream sequence in a buffer lane, for which the leftmost car is of the same color \citep[e.g.,][]{Leng.2020,Leng.2023}. We hence evaluate a fourth action mask $m^{\text{GS}}(s_t,a)$ given as

\begin{equation}
m^{\text{GS}}(s_t,a)=\begin{cases}
1\text{, if } a\in\{L+1,\dots,2L\} \text{ and } \exists j \in\{2,\dots,W\}\colon e_{t,1}=B_{t,a-L,j} \text{ and } B_{t,a-L,j-1}=0 \\
1 \text{, if } \nexists i \in\{1,\dots,L\}, j \in\{2,\dots,W\}\colon e_{t,1}=B_{t,i,j} \text{ and } B_{t,i,j-1}=0\\
0\text{, else}.
\end{cases}
\end{equation}
Although greedy storage presents a reasonable policy, we cannot prove that it is indeed optimal.

We also want to note that employing the aforementioned action masks does not fully solve the paint shop problem. In fact, greedy retrieval, greedy storage, and fast-track actions are often not applicable. So far, we proposed several action masks independently. However, as detailed in the following, we also combine and prioritize multiple action masks.

\subsubsection*{Combining action masks}

We consider two approaches of combining action masks. First, we define the combination $m_1 \oplus m_2$, which only allows an action $a$ to be performed in state $s_t$, if both masks allow $a$
\begin{equation}
(m_1 \oplus m_2)(s_t,a)=\begin{cases}
			1 \text{, if } m_1(s_t,a)=1 \land m_2(s_t,a)=1 \\
			0 \text{, else}.	
			\end{cases}
\end{equation}

The second combination $m_1 \ogreaterthan m_2$ applies two action masks in a sequential way, while assigning higher priority to $m_1$. This approach is particularly relevant when we want to prioritize one mask over another. We consider an action mask as applicable if it forbids at least one action. Hence, the combined mask $m_1 \ogreaterthan m_2$ applies $m_1$ if at least one action is not allowed by $m_1$, and $m_2$ otherwise
\begin{equation}
(m_1 \ogreaterthan m_2)(s_t,a)=\begin{cases}
			m_1(s_t,a) \text{, if } \exists a{:} \ m_1(s_t,a) = 0 \\
			m_2(s_t,a) \text{, else}.			
			\end{cases} 
\end{equation}
Following this notation, we evaluate the following combinations of action masks
\begin{itemize}
\item $m^{\text{INV}}\oplus  \left( \left(m^{\text{GR}} \ogreaterthan m^{\text{FT}} \right) \ogreaterthan m^{\text{GS}} \right)  $ (all action masks)
\item $m^{\text{INV}}\oplus \left(m^{\text{GR}} \ogreaterthan m^{\text{FT}} \right) $ (invalid + greedy retrieval + fast-track)
\item $m^{\text{INV}}\oplus m^{\text{GR}}$ (invalid + greedy retrieval)
\item $m^{\text{INV}}$ (invalid)
\end{itemize}

\subsection{Policy learning}
 
Reinforcement learning alternates between applying the current policy $\pi_\theta(a|s_t)$ to generate experience in the form of trajectories $(s_1, a_1, r_1),\dots, (s_{H},a_{H},r_{H})$ with $H$ denoting the size of the experience set and using this experience to perform policy updates by optimizing a loss function $L(\theta)$. 
We implement proximal policy optimization \citep[PPO,][]{schulman_proximal_2017} for policy learning as we found it to perform well for sequential decision problems \citep{Brammer.2022a,Brammer.2022b,Brammer.2021}, while being easy to implement and tune. PPO maintains a policy network $\pi_\theta(a \mid s_t)$ and a value function $V^\theta(s_t)$. The values $\pi_\theta(a \mid s_t)$ reflect the probabilities that the agent performs action $a$ in state $s_t$, while $V^\theta(s_t)$ estimates the expected return when starting from state $s_t$ and following the current policy. 
The policy parameters $\theta$ are given by the weights of the neural network.

We briefly summarize the core idea of PPO following \citet{schulman_proximal_2017}. 
Given experience $(s_1, a_1, r_1),\dots, (s_{H},a_{H},r_{H})$, PPO estimates the advantage as
\begin{equation}
\hat{A}_t = R_t - V^\theta(s_t), \quad R_t=\sum_{t'=t}^{H-1} r_{t'} .
\end{equation}
Policy learning maximizes the clipped surrogate objective
\begin{align}
\max_\theta\ L(\theta) = \sum\limits_{t=1}^{H} L_t^{CLIP}(\theta) - c_1 L_t^{VF}(\theta) + c_2 L_t^H(\theta).
\label{eqn:ppo_loss} 
\end{align}
with
\begin{align}
L_t^{CLIP}(\theta) = \min \left\lbrace \frac{\pi_\theta(a\,|\,s_t)}{\pi_{\theta_\mathrm{old}}(a\,|\,s_t)} \hat{A}_t, \text{clip} \left(\frac{\pi_\theta(a\,|\,s_t)}{\pi_{\theta_\mathrm{old}}(a\,|\,s_t)}, 1-\varepsilon, 1+ \varepsilon \right) \hat{A}_t \right\rbrace
\label{eqn:fraction_pi} 
\end{align}
Actions with positive advantage are assigned greater probabilities and vice versa.\footnote{Note that there are different methods for advantage estimation depending on the particular PPO implementation, e.g., generalized advantage estimation \citep{schulman_high-dimensional_2015}.} 
Clipping restricts the extent of policy updates, thereby stabilizing the learning process. 
The value loss $L_t^{\text{VF}}(\theta)=(R_t - V^\theta(s_t))^2$ improves the state-value estimate, while the entropy term
$L_t^{H}(\theta)= -\sum_{a\in A} \pi_\theta(a \,|\,s_t)\,\log_2 \pi_\theta(a\,|\,s_t)$ 
encourages sufficient exploration during training.

We implement our RL approach in Python 3.8.0 using the PPO implementation from the RL framework ``Stable Baselines 3'' \citep{stable-baselines3} in version 1.5.0. We train all models for a total of 10 million timesteps.\footnote{We conducted a convergence analysis which suggests that no further improvement is achieved when training longer. The results of the convergence analysis are presented in Appendix H.} Training a single policy on a machine with an AMD EPYC 7702 processor and 64\,GB main memory takes approximately ten hours. 
All parameters except the number of time steps are set to their default values as stated in Appendix A of the supplementary material.

\section{Evaluation}
\label{section-Evaluation}

\subsection{Procedure}

We perform a total of five analyses. First, the main analysis compares our RL approach against all baselines. The problem instances of all analyses (except the out-of-distribution instances) are generated from a distribution with equal probability for each color. The length of the upstream sequences is fixed to 100. We impose a cutoff time of five minutes for all approaches to model the necessity to quickly adapt to new incoming orders and short-term changes in production. Second, we compare our RL approach against multiple baselines with an extended cutoff time of ten hours. We expect Gurobi to solve small instances optimally, which allows us to assess the performance of our approach in comparison to the (near-)optimal Gurobi solutions. Third, we lift the assumption of square buffers and consider rectangular buffer sizes. Fourth, we consider problem instances with an initially filled buffer. While an initially empty buffer reflects the academic notion of the problem, real-world production never stands still. Hence, real-world instances of the paint shop problem also feature filled buffers. Fifth, we consider imbalanced (instead of balanced) color distributions. However, we deliberately do not retrain our RL approach to understand how it performs on out-of-distribution problems. Changes in the color distribution can, for instance, occur due to an unforeseen shortage of painting color and other supplies.


An overview of the considered problem instances is provided in \Cref{tbl:parameters_analysis}. We consider 5--15 colors, while the buffer size (lanes x lane width) ranges from 2x2 to 8x10. For all instances, we require the number of buffer lanes to be smaller than the number of colors to ensure that the problem does not become trivial. Our problem parameters are similar to those of existing studies, e.g., \citet{Leng.2020} and \citet{sun2017study}, who used 10, 14, and 20 colors and buffer sizes of 5x6, 7x8, and 10x10. Our problem instances are also publicly available, see \citet{psproblemdata}.

\begin{table}[htp]
	\caption{Problem parameters.} \label{tbl:parameters_analysis} 
	\centering
	\setlength{\tabcolsep}{1pt}
	\footnotesize
	\vspace{-0.2cm}
	\begin{tabular}{L{5cm}L{2.7cm}L{2.8cm}L{5.5cm}}
		\toprule
		Analysis & Sequence length & Number of colors & Buffer size (lanes x lane width) \\ \addlinespace
		\midrule
		Main analysis & 100 & 5, 10, 15 & 2x2, 3x3, 4x4, 5x5, 6x6, 7x7, 8x8\\
		Evaluation with 10h cutoff time & 100 & 5, 10, 15 & 2x2, 3x3, 4x4, 5x5, 6x6, 7x7, 8x8\\
		Rectangular buffers & 100 & 10 & 4x5, 4x6, 4x8, 5x6, 5x7, 5x8, 5x10, 6x4, 6x10, 7x8, 8x10 \\		
		Initially filled buffer & 100 & 5, 10, 15 & 2x2, 3x3, 4x4, 5x5, 6x6, 7x7, 8x8\\		
		Imbalanced color distribution & 100 & 5, 10, 15 & 2x2, 3x3, 4x4, 5x5, 6x6, 7x7, 8x8 \\

		\bottomrule

	\end{tabular}
\end{table}

\subsection{Baselines}
\label{section:Baselines}

We compare our RL approach against the following baselines: Gurobi based on our ILP formulation, Gurobi based on the store-then-retrieve formulation by \citet{wu2021mathematical}, the greedy heuristic, alternated greedy storage and RL for retrieval \citep{Huang.2024, Leng.2020, Leng.2023},  multiple sequence alignment (MSA), and variants of the metaheuristic simulated annealing.

\subsubsection*{Gurobi}

We implement our ILP formalization from \Cref{section-PS-problem} in Gurobi using ``gurobipy'' version 10.0.2.
Gurobi is a state-of-the-art commercial ILP solver, free for academic use, that employs optimized branch-and-bound algorithms. Implication constraints are also supported by the ``gurobipy'' package, which allows us to enter the constraints without prior normalization.

\subsubsection*{Gurobi based on store-then-retrieve problem formulation \citep{wu2021mathematical}}

We also evaluate Gurobi based on a modified version of the ILP formalization by \citet{wu2021mathematical}.  
We adapt their problem formulation to our problem in the following way: First, we remove car types to make all cars generic, i.e., of the same type. Second, we unify the costs of color changes to one as \citet{wu2021mathematical} allow the costs of color changes to differ, e.g., changing from red to blue might be more costly than changing from red to green. Third, we exclude the assembly shop and focus only on the paint shop.
Accordingly, we also place the buffer system in front of the paint shop instead. 
The adjusted ILP formulation is provided in Appendix I of the supplementary material. \citeauthor{wu2021mathematical}~consider the paradigm ``store-then-retrieve''. Accordingly, we iteratively repeat the ``store-then-retrieve'' procedure until the entire upstream sequence is processed.

\subsubsection*{Greedy}

We implement a greedy solution policy following the proposed action masks that enforce greedy retrieval, fast-track actions, and greedy storage. 
The greedy policy runs until the upstream sequence and buffer are both empty. Greedy alternates between storage and retrieval phases, while starting with the storage phase. Here, the heuristic performs greedy storage if possible. Otherwise, the current car is stored in the least occupied lane. In case of ties, the lane with the lowest index is used.  
Once the buffer is filled, greedy changes to the retrieval phase. The heuristic performs greedy retrieval if possible and retrieves from the lane with the lowest index otherwise. Accordingly the heuristic is fully deterministic, which is preferable in regard to reproducibility. Note that fast-track actions are implicitly performed by the proposed greedy heuristic since cars are stored in an empty lane in the storage phase. The next retrieve operation in the retrieval phase will perform the greedy retrieval. The pseudocode of Greedy is provided in Appendix J of the supplementary material.


Deciding when to switch from storage to retrieval presents a crucial challenge to the fully flexible problem version. Therefore, we evaluate another Greedy baseline that changes between storage and retrieval phase according to the buffer fill rate \citep[e.g.,][]{Huang.2024,sun2017study}. The fill rate is defined as the current number of cars in the buffer divided by the buffer size $LW$.
\REV{This Greedy approach has two parameters $F_u, F_\ell$ denoting the upper and lower bounds of the fill rate. In the evaluation, we consider all combinations $F_u, F_\ell \in \{0,0.01,0.02,\dots, 1\}$ and report the best respective result for each instance.}
A more fine-grained grid does not yield better results as for the largest buffer size of our dataset, i.e., 8x10, any difference of fill rates is already greater than $1/(8\cdot 10)>0.01$. The pseudocode of ``Greedy based on fill rate'' is provided in Appendix J of the supplementary material.

\subsubsection*{Alternated greedy storage and reinforcement learning for retrieval}

We consider another baseline that follows the RL approaches by \citet{Leng.2020}, \citet{Leng.2023}, and \citet{Huang.2024}. The buffer is first filled using greedy storage until it reaches the desired fill rate of 60\% \citep{Huang.2024,sun2017study}. Subsequently, the baseline alternates between inserting one car into the buffer with greedy storage and then retrieving one car with RL. Note that these studies focused on the problem variation with non-uniform costs of color changes and model types, whereas our problem has uniform model types and costs of color changes. Therefore, combining greedy storage with our own RL policy should yield a reasonable estimate of the performance of the existing RL approaches for the paint shop problem with multi-lane buffers.

\subsubsection*{Multiple sequence alignment}

We implement multiple sequence alignment (MSA), a dynamic programming approach inspired by bioinformatics and introduced by \citet{Epping.2003}. MSA is a retrieval-only algorithm that works by first ``merging'' consecutive cars of the same color, which will be retrieved together. MSA then finds an optimal alignment of buffer columns, so that as many cars in a column have the same color, and then retrieves them all together. In this way, MSA forms color batches across lanes and columns simultaneously. Given a buffer state, the retrieval sequence that MSA produces is optimal as shown by \citet{Epping.2003}. Although MSA only provides a retrieval policy, we combine it with greedy storage, to obtain a full storage and retrieval method. MSA has a time complexity of $ \mathcal{O}\left(2^L\prod_{i=1}^L W_i\right)$, where $W_i$ is the number of cars in buffer lane $i$. As a consequence, this approach requires longer than the cutoff time of five minutes to produce a solution for instances with 9x9 buffer.
Multiple sequence alignment is implemented in C++14.

\subsubsection*{Simulated annealing}

We implement the metaheuristic simulated annealing \citep[SA, e.g.,][]{gendreau2018handbook}.
SA improves an initial action sequence by iteratively replacing it with a new sequence from the neighborhood. If the new sequence reduces the number of color changes, the new sequence is directly accepted as the new solution. Otherwise, a sequence can still be accepted as the new solution depending on the current temperature. The temperature is higher in early iterations to allow for exploration but it cools down after each iteration. For this purpose, we implement geometric cooling ($T_{i} = \beta^i T_0$) \citep{goodson2015priori,knopp2017batch}. We determined the initial temperature $T_0$ and cooling factor $\beta$ through a pre-study, see Appendix E of the supplementary material.
We evaluate SA based on the initial action sequence generated by the greedy heuristic as it creates a feasible initial solution using the heuristics of greedy storage and retrieval.

Defining a neighborhood of feasible candidate solutions is challenging due to the structural constraints imposed on the action sequences. Arbitrary swaps or only updating one element will usually result in infeasible solutions.
We address this by leveraging the structure of the initial action sequence. We first identify the switch points where the action changes between storage and retrieval. Thereby, we partition the sequence into different segments of storage or retrieval operations. We define the neighborhood as all possible permutations of the operations within each segment. This ensures feasibility and allows for meaningful local perturbations. By performing repeated pairwise swaps, we generate arbitrary permutations in each segment which enables effective exploration of the solution space without violating the problem's constraints. The pseudocode of SA and an example of a neighborhood sequence are provided in Appendix D of the supplementary material. Simulated annealing is implemented in C++14.

\subsection{Application of a trained reinforcement learning policy}

The trained policy outputs a probability distribution over all actions for a given state. Based on this, we apply a trained policy in two ways. First, we apply the policy in a deterministic manner by always selecting the action with the highest probability. Second, we use the available cutoff time to continuously sample actions from the policy to search for the best solution. Thereby, we may find superior solutions than those of the deterministic application. In all analyses, we ignore the time needed for policy learning as this can be performed in advance for a given paint shop.

\subsection{Performance metrics}
\label{section:Performance}

Our main performance metric is given by the \emph{Average Relative Percentage Deviation} \citep[ARPD, e.g.,][]{Brammer.2022a,Huang.2024}. To define ARPD, we first define the relative percentage deviation (RPD). Let $c^{i,A}=\sum_{t=1}^{2N-1} z_t$ denote the number of color changes of approach $A$ on problem instance $i$. In addition, let $c^{i,A^*}$ denote the total number of color changes of the best solution on instance $i$ that was found by some algorithm $A^*$. $RPD^{i,A}$ is then defined as
\begin{align}
\text{RPD}^{i,A}=\frac{c^{i,A}  - c^{i,A^*}}{ c^{i,A^*} } \cdot 100.
\label{eqn:rpd}
\end{align}
Note that $c^{i,A^*} \neq 0$ for every problem with more than one color. Now, let $\mathcal{I}$ denote the set of all considered problem instances. The ARPD of approach $A$ is given as
\begin{align}
\text{ARPD}^A=\frac{ \sum_{i \in \mathcal{I}} \text{RPD}^{i,A} }{|\mathcal{I}|}.
\label{eqn:arpd}
\end{align}
If $\text{ARPD}^A =3.20$, the downstream sequences generated by approach $A$ have on average \SI{3.20}{\percent} more color changes than the best solution that was found by some approach. Note that the best relative solution is not necessarily an optimal solution. 

The metric ARPD is well suited to compare the relative performance of different approaches but it does not provide any information regarding the absolute number of color changes. Therefore, we also report the mean number of color changes caused by approach $A$
\begin{equation}
c^{A}=\frac{ \sum_{i \in \mathcal{I}}  \sum_{t=1}^{2N-1} z_t }{|\mathcal{I}|}.
\end{equation}

\section{Results}
\label{section-Results}

\subsection{Main analysis}


We first consider the results of our main analysis, which compares the performance of all approaches on problem instances with quadratic buffer and balanced color distribution. The results are shown in \Cref{tbl:quadratic}. Due to spatial limitations, we included only a subset of the results in \Cref{tbl:quadratic} and excluded buffer sizes 3x3, 5x5, and 7x7 for 10 and 15 colors. The remaining results and computation times are provided in Appendices F and G of the supplementary material. 


We find that our RL approach with all action masks and repeated sampling performs best on most problem instances. In fact, repeated sampling from the RL policy always outperforms its respective counterparts of RL with deterministic policy application. However, on the problems with 15 colors and 2x2 buffer, Gurobi achieves the best performance as they are comparably easier to solve due to the small buffer size.

\begin{table}[H]
\setlength{\tabcolsep}{2.5pt}
\caption{Evaluation results (ARPD) for main analysis. \label{tbl:quadratic}}

\footnotesize
\centering
\vspace{-0.2cm}
\begin{tabular}{l*{11}{S[detect-weight, mode=text,table-format=2.1,table-column-width=0.8cm]}}
\toprule

Colors & \multicolumn{3}{c}{5} & \multicolumn{4}{c}{10} & \multicolumn{4}{c}{15} \\
 \cmidrule(lr){2-4} \cmidrule(lr){5-8} \cmidrule(lr){9-12}
Buffer size (lanes x lane width) &    {2x2} &     {3x3} &     {4x4} &     {2x2} &   {4x4} &   {6x6} &   {8x8}  &    {2x2} &   {4x4} &   {6x6} &   {8x8} \\

\midrule

\multicolumn{11}{l}{\underline{Reinforcement learning with stochastic policy application (sampling)}} \\ \addlinespace 
\;\;All action masks & 2.9 & \B 0.6 & \B 0.0 & 4.7 & \B 0.0 & \B  0.7 & 13.8 & 4.5 & \B 0.5 & \B  4.4 & 25.2 \\
\;\;Invalid + greedy retrieval + fast-track  & 6.5 & 5.1 & 9.2 & 5.7 & 9.0 & 43.1 & 79.3 & 4.5 & 8.5 & 31.3 & 61.0 \\
\;\;Invalid + greedy retrieval & 5.2 & 9.8 & 22.0 & 7.9 & 16.6 & 43.6 & 99.3 & 4.9 & 14.6 & 40.1 & 77.1 \\
\;\;Invalid & 8.3 & 17.1 & 34.8 & 6.7 & 33.1 & 57.1 & 108.8 & 8.3 & 24.8 & 53.7 & 107.2 \\
\;\;No action masks & \B 1.5 & 20.7 & 22.5 & \B 2.2 & 25.6 & 57.7 & 107.8 & 2.6 & 24.9 & 87.5 & 174.2 \\

\addlinespace

\multicolumn{11}{l}{\underline{Reinforcement learning with deterministic policy application}} \\ \addlinespace

\;\;All action masks  & 10.1 & 15.4 & 24.4 & 10.9 & 26.4 & 33.7 & 62.1 & 10.2 & 19.4 & 32.2 & 55.2 \\
\;\;Invalid + greedy retrieval + fast-track & 12.7 & 22.0 & 35.4 & 13.9 & 34.7 & 76.4 & 119.0 & 12.0 & 26.3 & 53.9 & 96.4 \\
\;\;Invalid + greedy retrieval & 15.6 & 30.8 & 40.6 & 15.2 & 36.5 & 73.8 & 150.6 & 12.4 & 34.6 & 64.0 & 110.3 \\
\;\;Invalid & 17.2 & 37.4 & 53.2 & 13.2 & 47.1 & 81.3 & 149.0 & 15.3 & 37.3 & 71.3 & 130.8 \\
\;\;No action masks  & 9.6 & 39.6 & 49.0 & 12.4 & 40.5 & 93.8 & 156.8 & 12.0 & 43.5 & 106.2 & 201.3 \\

\addlinespace
\multicolumn{11}{l}{\underline{Baselines}} \\ \addlinespace

\;\;Gurobi (our ILP formulation) & 15.9 & 99.4 & 206.9 & 9.9 & 69.9 & 170.4 & 341.7 & \B 2.3 & 41.1 & 107.5 & 196.9 \\
\;\;Gurobi \citep{wu2021mathematical} & 32.2 & 32.4 & 28.5 & 19.3 & 9.3 & 76.1 & 233.8 & 12.8 & 4.8 & 45.5 & 129.8 \\
\;\;Greedy & 33.6 & 72.4 & 101.5 & 23.3 & 46.5 & 62.7 & 80.5 & 15.7 & 38.6 & 50.4 & 55.2 \\
\;\;Greedy based on fill rate & 25.0 & 39.4 & 35.9 & 16.3 & 19.7 & 7.8 & \B 2.6 & 12.8 & 12.5 & 12.1 & 13.6 \\
\;\;Alternate greedy storage and RL & 19.8 & 33.2 & 37.3 & 13.5 & 28.5 & 17.7 & 18.0 & 13.4 & 15.1 & 24.8 & 19.9 \\
\;\;Multiple sequence alignment & 40.6 & 54.3 & 67.8 & 25.7 & 24.6 & 15.1 & 13.8 & 19.6 & 14.8 & 10.1 & 16.6 \\
\;\;Simulated annealing & 32.7 & 51.0 & 65.8 & 20.6 & 17.4 & 13.0 & 20.1 & 15.4 & 20.5 & 5.5 & \B 2.7 \\

\bottomrule
\end{tabular}

\end{table}

\REV{On the problems with 10 or 15 colors and 8x8 buffers, Greedy based on fill rate and SA present the best performing approaches respectively. In particular, this Greedy method consistently outperforms the ``store-then-retrieve''  Greedy approach, which fills the buffer first using greedy storage to empty it later using greedy retrieval, if applicable. SA achieves the lowest ARPD on problems with 15 colors and 8x8 buffer. 
A possible explanation could be that RL struggles to sufficiently explore the considerably larger state space, caused by the increased problem complexity.

We perform a Friedman test \citep{Friedman.1937}, followed by pairwise Nemenyi post-hoc tests \citep{Nemenyi.1963}, to assess whether the performance differences among the three approaches (RL, Greedy based on fill rate, simulated annealing)  are statistically significant. We use the instance size as blocking factor in the randomized block design analysis. The overall Friedman statistic is $Q(2) = 16.2,\ p < 0.001$. The pairwise Nemenyi comparisons indicate significant differences for \emph{RL vs.\ Greedy based on fill rate} ($p < 0.05$) and \emph{RL vs.\ Simulated annealing} ($p < 0.01$).

}


\definecolor{bronze}{rgb}{0.8, 0.5, 0.2}
\definecolor{auburn}{rgb}{0.43, 0.21, 0.1}

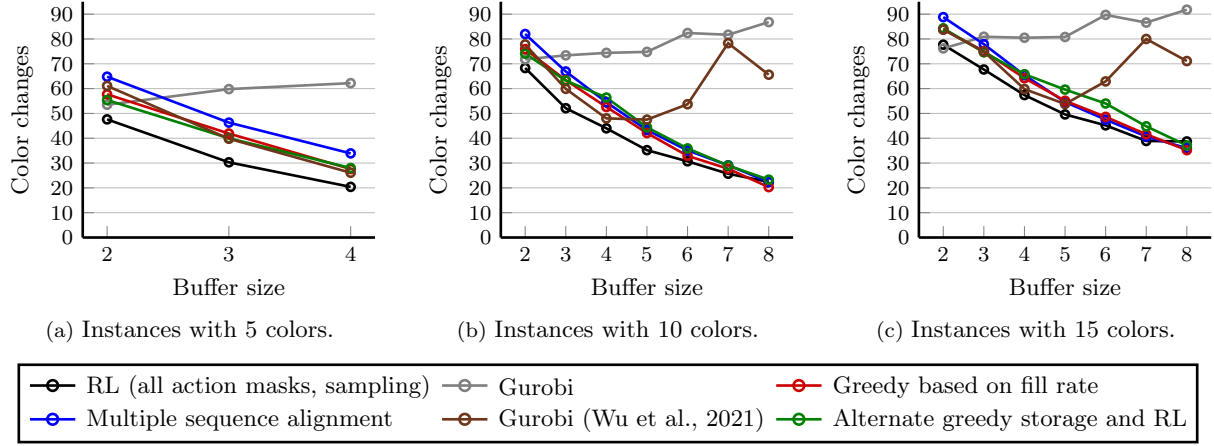
\begin{figure}[H]
\footnotesize
\begin{subfigure}[b]{0.33\textwidth}
\centering
\pgfplotsset{every tick label/.append style={font=\scriptsize}}
     \centering
        \begin{tikzpicture}
		\begin{axis}[
			width=\textwidth,
		y label style={font=\footnotesize},
		x label style={font=\footnotesize},
		title style={font=\small},
		xtick={2, 3, 4,5,6,7,8},
		ytick={0, 10,20, 30, 40,50,60,70,80, 90},
		ymax=95, ymin=0,
		  ylabel={Color changes},
	 		xlabel={Buffer size},
		   ymajorgrids,
		   axis lines*=left,
		  mark size=1.5pt,
		  line width=1pt,
		   legend style={
				anchor=north,
				font=\tiny,
				legend cell align=left,
				legend pos=south west,
				draw = none
			}
		]

		\addplot[color=black,mark = o, line width=1pt] table [x=Buffer size, y=RL with AM (rep), col sep=comma] {regular_5colors.csv}; \label{pgfplots:main_RL}
		\addplot[color=gray,mark = o, line width=1pt] table [x=Buffer size, y=Gurobi, col sep=comma] {regular_5colors.csv}; \label{pgfplots:main_gurobi}
		\addplot[color=blue,mark = o, line width=1pt] table [x=Buffer size, y=MSA, col sep=comma] {regular_5colors.csv}; \label{pgfplots:main_msa}
		\addplot[color=red,mark = o, line width=1pt] table [x=Buffer size, y=Greedy_fillrate, col sep=comma] {regular_5colors.csv}; \label{pgfplots:main_greedy_fillrate}
		\addplot[color=green,mark = o, line width=1pt] table [x=Buffer size, y=GSplusRL, col sep=comma] {regular_5colors.csv}; \label{pgfplots:main_GSplusRL}
		\addplot[color=auburn,mark = o, line width=1pt] table [x=Buffer size, y=Gurobi (Wu), col sep=comma] {regular_5colors.csv}; \label{pgfplots:main_Gurobi_Wu}
		\end{axis}
		\end{tikzpicture}
		\caption{\footnotesize Instances with 5 colors.}
\end{subfigure}
\begin{subfigure}[b]{0.33\textwidth}
\centering
\pgfplotsset{every tick label/.append style={font=\scriptsize}}
     \centering
        \begin{tikzpicture}
		\begin{axis}[
			width=\textwidth,
		y label style={font=\footnotesize},
		x label style={font=\footnotesize},
		title style={font=\small},
		xtick={2, 3, 4,5,6,7,8},
		ytick={0, 10,20, 30, 40,50,60,70,80, 90},
		ymax=95, ymin=0,
		  ylabel={Color changes},
	 		xlabel={Buffer size},
		   ymajorgrids,
		   axis lines*=left,
		  mark size=1.5pt,
		  line width=1pt,
		   legend style={
				anchor=north,
				font=\tiny,
				legend cell align=left,
				legend pos=south west,
				draw = none
			}
		]
		\addplot[color=black,mark = o, line width=1pt] table [x=Buffer size, y=RL with AM (rep), col sep=comma] {regular_10colors.csv}; \label{pgfplots:main_RL}
		\addplot[color=gray,mark = o, line width=1pt] table [x=Buffer size, y=Gurobi, col sep=comma] {regular_10colors.csv}; \label{pgfplots:main_gurobi}
		\addplot[color=blue,mark = o, line width=1pt] table [x=Buffer size, y=MSA, col sep=comma] {regular_10colors.csv}; \label{pgfplots:main_msa}
		\addplot[color=red,mark = o, line width=1pt] table [x=Buffer size, y=Greedy_fillrate, col sep=comma] {regular_10colors.csv}; \label{pgfplots:main_greedy_fillrate}
		\addplot[color=green,mark = o, line width=1pt] table [x=Buffer size, y=GSplusRL, col sep=comma] {regular_10colors.csv}; \label{pgfplots:main_GSplusRL}
		\addplot[color=auburn,mark = o, line width=1pt] table [x=Buffer size, y=Gurobi (Wu), col sep=comma] {regular_10colors.csv}; \label{pgfplots:main_Gurobi_Wu}
		\end{axis}
		\end{tikzpicture}
		\caption{\footnotesize Instances with 10 colors.}
\end{subfigure}
\begin{subfigure}[b]{0.33\textwidth}
\centering
\pgfplotsset{every tick label/.append style={font=\scriptsize}}
     \centering
        \begin{tikzpicture}
		\begin{axis}[
			width=\textwidth,
		y label style={font=\footnotesize},
		x label style={font=\footnotesize},
		title style={font=\small},
		xtick={2, 3, 4,5,6,7,8},
		ytick={0, 10,20, 30, 40,50,60,70,80, 90},
		ymax=95, ymin=0,
		  ylabel={Color changes},
	 		xlabel={Buffer size},
		   ymajorgrids,
		   axis lines*=left,
		  mark size=1.5pt,
		  line width=1pt,
		   legend style={
				anchor=north,
				font=\tiny,
				legend cell align=left,
				legend pos=south west,
				draw = none
			}
		]
		\addplot[color=black,mark = o, line width=1pt] table [x=Buffer size, y=RL with AM (rep), col sep=comma] {regular_15colors.csv}; \label{pgfplots:main_RL}
		\addplot[color=gray,mark = o, line width=1pt] table [x=Buffer size, y=Gurobi, col sep=comma] {regular_15colors.csv}; \label{pgfplots:main_gurobi}
		\addplot[color=blue,mark = o, line width=1pt] table [x=Buffer size, y=MSA, col sep=comma] {regular_15colors.csv}; \label{pgfplots:main_msa}
		\addplot[color=red,mark = o, line width=1pt] table [x=Buffer size, y=Greedy_fillrate, col sep=comma] {regular_15colors.csv}; \label{pgfplots:main_greedy_fillrate}
		\addplot[color=green,mark = o, line width=1pt] table [x=Buffer size, y=GSplusRL, col sep=comma] {regular_15colors.csv}; \label{pgfplots:main_GSplusRL}
		\addplot[color=auburn,mark = o, line width=1pt] table [x=Buffer size, y=Gurobi (Wu), col sep=comma] {regular_15colors.csv}; \label{pgfplots:main_Gurobi_Wu}
		\end{axis}
		\end{tikzpicture}
		\caption{\footnotesize Instances with 15 colors.}
\end{subfigure}

\vspace{0.2cm}
\centering
\begin{tikzpicture}
	\matrix[
	 font=\footnotesize,
    matrix of nodes,
    anchor=west,
    draw,
    inner sep=0.2em,
    line width=1pt,
    cells={line width=1pt, anchor=west}
  ]
  at(0.6,2.2){
    \ref{pgfplots:main_RL} & RL (all action masks, sampling) &
    \ref{pgfplots:main_gurobi} & Gurobi &
    \ref{pgfplots:main_greedy_fillrate} & Greedy based on fill rate 
     \\
    \ref{pgfplots:main_msa} & Multiple sequence alignment &
    \ref{pgfplots:main_Gurobi_Wu} & Gurobi \citep{wu2021mathematical} &
    \ref{pgfplots:main_GSplusRL} & Alternate greedy storage and RL \\
   };
    \end{tikzpicture}
\caption{Evaluation results (color changes) of main analysis for best RL approach and baselines.\label{fig:main_results_plot}}
\end{figure}

\definecolor{armygreen}{rgb}{0.29, 0.33, 0.13}
\definecolor{antiquefuchsia}{rgb}{0.57, 0.36, 0.51}

\begin{figure}[H]
\footnotesize
\begin{subfigure}[b]{0.33\textwidth}
\pgfplotsset{every tick label/.append style={font=\scriptsize}}
     \centering
     \begin{tikzpicture}
		\begin{axis}[
			width=\textwidth,
		y label style={font=\footnotesize},
		x label style={font=\footnotesize},
		xtick={2, 3, 4,5,6,7,8},
		ytick={0, 10,20, 30, 40,50,60,70,80, 90},
		ymax=95, ymin=0,
		  ylabel={Color changes},
	 		xlabel={Buffer size},
		   ymajorgrids,
		   axis lines*=left,
		  mark size=1.5pt,
		  line width=1pt,
		   legend style={
				anchor=north,
				legend columns=3,
				font=\footnotesize,
				legend cell align=left,
				at={(0.5,-0.25)},
			}
		]
		\addplot[color=black,mark = o, line width=1pt] table [x=buffer_size, y=Max, col sep=comma] {ps_5colors.csv};\label{pgfplots:AM_sampling}
		\addplot[color=armygreen,mark = o, line width=1pt] table [x=buffer_size, y=FT_rep, col sep=comma] {ps_5colors.csv};\label{pgfplots:AM_det}
		\addplot[color=antiquefuchsia,mark = o, line width=1pt] table [x=buffer_size, y=no_mask_repeated, col sep=comma] {ps_5colors.csv};\label{pgfplots:NoAM_sampling}
		\addplot[color=bronze,mark = o, line width=1pt] table [x=buffer_size, y=PPO_vanilla, col sep=comma] {ps_5colors.csv};\label{pgfplots:NoAM_det}
		\end{axis}
		\end{tikzpicture}
		\caption{\footnotesize Instances with 5 colors.}
\end{subfigure}	
\begin{subfigure}[b]{0.33\textwidth}
\pgfplotsset{every tick label/.append style={font=\scriptsize}}
     \centering
     \begin{tikzpicture}
		\begin{axis}[
			width=\textwidth,
		y label style={font=\footnotesize},
		x label style={font=\footnotesize},
		xtick={2, 3, 4,5,6,7,8},
		ytick={0, 10,20, 30, 40,50,60,70,80, 90},
		ymax=95, ymin=0,
		  ylabel={Color changes},
	 		xlabel={Buffer size},
		   ymajorgrids,
		   axis lines*=left,
		  mark size=1.5pt,
		  line width=1pt,
		   legend style={
				anchor=north,
				legend columns=3,
				font=\footnotesize,
				legend cell align=left,
				at={(0.5,-0.25)},
			}
		]
		\addplot[color=black,mark = o, line width=1pt] table [x=buffer_size, y=Max, col sep=comma] {ps_10colors.csv};\label{pgfplots:AM_sampling}
		\addplot[color=armygreen,mark = o, line width=1pt] table [x=buffer_size, y=FT_rep, col sep=comma] {ps_10colors.csv};\label{pgfplots:AM_det}
		\addplot[color=antiquefuchsia,mark = o, line width=1pt] table [x=buffer_size, y=no_mask_repeated, col sep=comma] {ps_10colors.csv};\label{pgfplots:NoAM_sampling}
		\addplot[color=bronze,mark = o, line width=1pt] table [x=buffer_size, y=PPO_vanilla, col sep=comma] {ps_10colors.csv};\label{pgfplots:NoAM_det}
		\end{axis}
		\end{tikzpicture}
		\caption{\footnotesize Instances with 10 colors.}
\end{subfigure}	
\begin{subfigure}[b]{0.33\textwidth}
\pgfplotsset{every tick label/.append style={font=\scriptsize}}
     \centering
     \begin{tikzpicture}
		\begin{axis}[
			width=\textwidth,
		y label style={font=\footnotesize},
		x label style={font=\footnotesize},
		xtick={2, 3, 4,5,6,7,8},
		ytick={0, 10,20, 30, 40,50,60,70,80, 90},
		ymax=95, ymin=0,
		  ylabel={Color changes},
	 		xlabel={Buffer size},
		   ymajorgrids,
		   axis lines*=left,
		  mark size=1.5pt,
		  line width=1pt,
		   legend style={
				anchor=north,
				legend columns=3,
				font=\footnotesize,
				legend cell align=left,
				at={(0.5,-0.25)},
			}
		]
		\addplot[color=black,mark = o, line width=1pt] table [x=buffer_size, y=Max, col sep=comma] {ps_15colors.csv};\label{pgfplots:AM_sampling}
		\addplot[color=armygreen,mark = o, line width=1pt] table [x=buffer_size, y=FT_rep, col sep=comma] {ps_15colors.csv};\label{pgfplots:AM_det}
		\addplot[color=antiquefuchsia,mark = o, line width=1pt] table [x=buffer_size, y=no_mask_repeated, col sep=comma] {ps_15colors.csv};\label{pgfplots:NoAM_sampling}
		\addplot[color=bronze,mark = o, line width=1pt] table [x=buffer_size, y=PPO_vanilla, col sep=comma] {ps_15colors.csv};\label{pgfplots:NoAM_det}
		\end{axis}
		\end{tikzpicture}
		\caption{\footnotesize Instances with 15 colors.}
\end{subfigure} 	

\vspace{0.2cm}

\centering

\begin{tikzpicture}
	\matrix[
	 font=\footnotesize,
    matrix of nodes,
    anchor=west,
    draw,
    inner sep=0.2em,
    line width=1pt,
    cells={line width=1pt, anchor=west}
  ]
  at(0.6,2.2){
    \ref{pgfplots:AM_sampling} & RL (all action masks, sampling) &
    \ref{pgfplots:AM_det} & RL (all action masks, deterministic) \\
    \ref{pgfplots:NoAM_sampling} & RL (no action masks, sampling) &
    \ref{pgfplots:NoAM_det} & RL (no action masks, deterministic) \\
   };
    \end{tikzpicture}

        \caption{Evaluation results (color changes) of main analysis for different RL policy applications.}
        
        \label{fig:actionmasking_comparison_plots}
\end{figure}
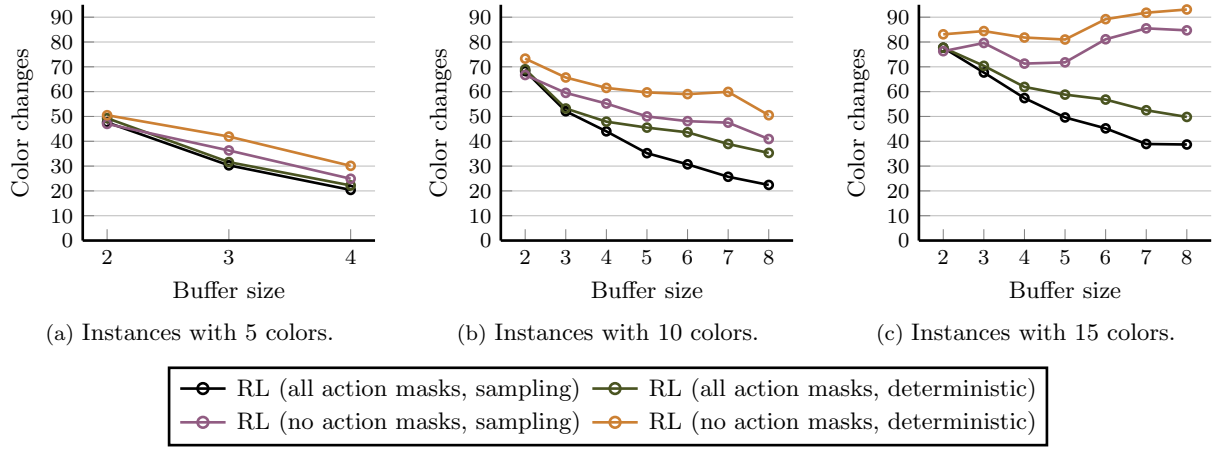


Applying Gurobi to different ILP formulations does not lead to competitive results. Neither ILP problem formulation (ours vs \citet{wu2021mathematical}), is superior to the other. Gurobi based on our problem formulation performs well on problems with 2x2 buffer, whereas Gurobi based on \citet{wu2021mathematical} achieves its lowest ARPD on problems with 15 colors and 4x4, 5x5 buffer.\footnote{See the complete results in Appendix F of the supplementary material.} Finally, we observe that following the paradigm of alternating between greedy storage and RL for retrieval at each timestep while maintaining a buffer fill rate of 60 \% also does not yield competitive results. Multiple sequence alignment performs poorly on the problem instances with five colors but its relative performance increases on larger problem instances with 10 and 15 colors.


\Cref{fig:main_results_plot} visualizes the mean number of color changes over ten instances per parameter setting for RL with all action masks and policy sampling against several baselines. Evidently, RL with all action masks and policy sampling usually achieves the lowest number of color changes. However, for problems with 8x8 buffer, RL is outperformed by Greedy based on fill rate. In addition, we observe that the number of color changes decreases for larger buffers, which is expected. Yet, this finding does not hold for the Gurobi baselines, which can be explained by the increased problem complexity due to greater buffer sizes.


We also analyze the influence of different combinations of action masks and policy applications as shown in \Cref{fig:actionmasking_comparison_plots}. Across all color settings, we find that including more action masks improves the overall performance for deterministic and repeated stochastic policy application. In addition, we observe that the improvement of  action masks and repeated stochastic policy application depends on the buffer size. While the benefit is rather small for 2x2 buffers, the performance increase of action masking is maximized for large buffers. For 10 colors and 8x8 buffers, including action masks and sampling from the policy reduces color changes by approximately 50\% compared to deterministic policy application without action masking. This improvement is even greater for 15 colors and 8x8 buffers. Furthermore, we find that, from no action masks and deterministic policy application, the performance benefit of including all action masks dominates the effect of repeated policy sampling. Accordingly, we only focus on RL with all action masks and repeated policy sampling for all subsequent analyses.

\subsection{Benchmarks with ten hours time limit}


We now compare the solutions of our RL approach with the possible (near-)optimal solutions found by several competing approaches with a time limit of ten hours. Here, we evaluate Gurobi based on our problem formulation, Gurobi based on the ``store-then-retrieve'' problem studied by \citet{wu2021mathematical}, and simulated annealing. Given the high time limit, Gurobi should be able to solve at least several of the small instances optimally. To account for the increased time limit of simulated annealing, we adjusted the cooling schedule to ensure that the final temperature for ten hours equals the final temperature after 5 minutes. We achieved this by rescaling the time index for the ten hour computation with the factor $\frac{10\text{h}}{5\text{min}}=120$. We specifically kept the sampling time of RL at five minutes like in our main analysis.


The results are provided in \Cref{tbl:Gurobi10h}. As expected, we find that Gurobi (based on our ILP formulation) is dominant for the small problem instances with 2x2 buffer. Here, the number of color changes of the RL solutions differ on average by 5.3--11.9\% from the near-optimal Gurobi solutions. However, even with a time limit of ten hours, Gurobi never managed to find an optimal solution of the 2x2 problems. For the larger problems with 3x3 buffer and larger, Gurobi (10h) is no longer competitive to RL with all action masks. Nevertheless, the extended cutoff time for Gurobi still leads to considerable reductions in ARPD compared to five minute cutoff time.

\begin{table}[H]
\setlength{\tabcolsep}{2.6pt}
\caption{Evaluation results (ARPD) with time limit 10h. \label{tbl:Gurobi10h}}
\footnotesize
\centering
\vspace{-0.2cm}
\begin{tabular}{l*{11}{S[detect-weight, mode=text,table-format=2.1,table-column-width=0.8cm]}}
\toprule
Colors & \multicolumn{3}{c}{5} & \multicolumn{4}{c}{10} & \multicolumn{4}{c}{15} \\
 \cmidrule(lr){2-4} \cmidrule(lr){5-8} \cmidrule(lr){9-12}
Buffer size (lanes x lane width) &    {2x2} &     {3x3} &     {4x4} &     {2x2} &   {3x3} &   {4x4} &   {8x8}  &    {2x2} &   {3x3} &   {4x4} &   {8x8} \\
\midrule

RL all action masks (sampling, 5min) & 5.3 & \B 0.0 & \B 0.0 & 11.0 & \B 0.0 & \B 0.0 & 22.6 & 11.9 & \B 0.6 & \B 0.5 & 66.2 \\ \addlinespace

Gurobi (our ILP formulation, 5min) & 18.1 & 98.1 & 206.9 & 16.5 & 41.6 & 69.9 & 373.9 & 9.5 & 20.3 & 41.1 & 294.0 \\ 
Gurobi (our ILP formulation, 10h)       & \B 0.2 & 52.1 & 107.0 & \B 0.0 & 30.4 & 51.9 & 248.5 & \B 0.0 & 9.6 & 23.7 & 214.1 \\ \addlinespace

Gurobi \citep[5min]{wu2021mathematical} & 35.3 & 31.6 & 28.5 & 26.5 & 15.1 & 9.3 & 258.3 & 20.7 & 11.4 & 4.8 & 205.0 \\ 
Gurobi \citep[10h]{wu2021mathematical} & 35.3 & 31.6 & 28.0 & 26.5 & 15.1 & 8.1 & 76.3 & 20.7 & 11.4 & 4.8 & 81.7 \\ \addlinespace

Simulated annealing (5min) & 36.0 & 50.2 & 65.2 & 27.9 & 29.3 & 17.4 & 28.9 & 23.3 & 18.8 & 20.7 & 36.5 \\
Simulated annealing (10h) & 33.5 & 46.5 & 60.4 & 24.8 & 25.4 & 11.9 & \B 0.5 & 21.1 & 12.9 & 13.3 & \B 0.0 \\

\bottomrule
\end{tabular}
\end{table}


Gurobi based on the ILP formulation by \citet{wu2021mathematical} does not benefit from 10h time limit for smaller instances as the 5min cutoff time is already sufficient. However, we observe considerable improvements for larger buffers as Gurobi \citep[10h]{wu2021mathematical} outperforms all approaches for problems with 5x5 buffer and 10 or 15 colors, see Appendix F of the supplementary material. 

Concerning simulated annealing, we observe strong improvements when increasing the cutoff time from five minutes to ten hours, in particular for larger problem sizes. While RL remains superior for problems with smaller buffers, simulated annealing (10h) outperforms all other approaches on instances with 10 or 15 colors and 6x6, 7x7, or 8x8 buffers.

\subsection{Rectangular buffers}

Next, we consider rectangular buffers instead of quadratic buffers. For this analysis, we fixed the number of colors to 10 and chose a selection of rectangular buffer sizes that were considered in prior studies  \citep{Huang.2024,Lin.2011,spieckermann_sequential_2004,sun2017study}. For simulated annealing, we used the optimal parameters based on the number of lanes. The results are shown in \Cref{tbl:rectangular_buffer}.
We find that RL still yields the best results out of all approaches for most buffer sizes. However, Greedy based on fill rate presents the best approach for 7x8 and 8x10 buffers.
In summary, the results suggest that the performance of our RL approach is not limited to quadratic buffers; instead, it also persists on rectangular buffers. \REV{The Friedman and Nemenyi tests confirm this observation, see Appendix K of the supplementary material.}

\begin{table}[H]
\setlength{\tabcolsep}{2.5pt}
\caption{Evaluation results (ARPD) for rectangular buffers.}
\label{tbl:rectangular_buffer}

\footnotesize
\centering
\vspace{-0.2cm}
\begin{tabular}{l*{11}{S[detect-weight, mode=text,table-format=2.1,table-column-width=0.8cm]}}
\toprule

Buffer size (lanes x lane width) &  {4x5} &  {4x6} &  {4x8} &  {5x6}&  {5x7}&  {5x8}&  {5x10} &  {6x4}&  {6x10} &  {7x8} &  {8x10} \\

\midrule
RL all action masks (sampling)       & \B 1.2 & \B 2.1 & \B 8.6 & \B 3.7 &\B 1.3 & \B 2.8 & \B 0.3 & \B 0.3 & \B 3.9 & 5.1 & 17.2 \\

\addlinespace
\multicolumn{11}{l}{\underline{Baselines}} \\ \addlinespace

\;\;Gurobi (our ILP formulation)  & 81.3 & 104.1 & 123.4 & 134.7 & 142.4 & 156.2 & 144.2 & 120.0 & 226.6 & 213.2 & 392.2 \\
\;\;Gurobi \citep{wu2021mathematical} & 50.6 & 55.6 & 64.4 & 73.5 & 75.5 & 63.4 & 63.1 & 70.6 & 61.2 & 72.8 & 61.1 \\
\;\;Greedy & 50.6 & 55.6 & 64.4 & 73.5 & 75.5 & 63.4 & 63.1 & 70.6 & 61.2 & 72.8 & 61.1 \\
\;\;Greedy based on fill rate & 18.8 & 15.0 & 22.6 & 20.6 & 16.0 & 15.4 & 10.7 & 19.6 & 4.1 & \B 4.2 &\B 3.2 \\
\;\;Alternate greedy storage and RL & 21.6 & 33.8 & 33.3 & 34.0 & 21.1 & 30.0 & 29.5 & 17.7 & 19.4 & 17.7 & 18.2 \\
\;\;Multiple sequence alignment & 23.5 & 23.3 & 24.8 & 30.9 & 16.6 & 21.9 & 6.7 & 29.1 & 12.2 & 9.0 & 28.1 \\
\;\;\REV{Simulated annealing} & 10.2 & 7.7 & 11.0 & 13.3 & 11.6 & 9.7 & 61.3 & 21.9 & 60.5 & 15.9 & 32.4 \\

\bottomrule
\end{tabular}
\end{table}

\subsection{Buffer initially filled}

We also consider problem instances, where the buffer is not empty but initially filled. Thereby, the problem setting is shifted from the academic notion towards a real-world scenario. We created new problem instances with a random buffer content. Note that these instances have higher complexity as the number of required retrieve operations increases from 100 to $100+L\times W$. We did not retrain the RL policies based on these novel instances to ensure a fair comparison.

\begin{table}[H]
\setlength{\tabcolsep}{3pt}
\caption{Evaluation results (ARPD) for initially filled buffer.}
\label{tbl:filled_buffer}

\footnotesize
\centering
\vspace{-0.2cm}
\begin{tabular}{l*{11}{S[detect-weight, mode=text,table-format=2.1,table-column-width=0.8cm]}}
\toprule
Colors & \multicolumn{3}{c}{5} & \multicolumn{4}{c}{10} & \multicolumn{4}{c}{15} \\
 \cmidrule(lr){2-4} \cmidrule(lr){5-8} \cmidrule(lr){9-12}
Buffer size (lanes x lane width) &    {2x2} &     {3x3} &     {4x4} &     {2x2} &   {4x4} &   {6x6} &   {8x8}  &    {2x2} &   {4x4} &   {6x6} &   {8x8} \\
\midrule
RL all action masks (sampling)        & \B 0.0 & \B 0.0 & \B 0.0 & \B 1.8 & \B 0.0 & \B 0.8 & \B 0.0 & \B 1.7 & \B 0.3 & \B 2.0 & 7.2 \\

\addlinespace
\multicolumn{11}{l}{\underline{Baselines}} \\ \addlinespace

\;\;Gurobi (our ILP formulation) & 10.7 & 93.1 & 158.1 & 5.2 & 71.3 & 123.3 & \textendash$^{\rm{a}}$ & 4.5 & 40.6 & 79.1 & 109.7 \\
\;\;Gurobi \citep{wu2021mathematical} & 22.4 & 36.3 & 55.5 & 15.0 & 20.4 & 99.7 & 162.6 & 9.8 & 10.9 & 58.6 & 105.8 \\
\;\;Greedy & 26.2 & 58.5 & 77.4 & 18.6 & 44.0 & 61.3 & 69.7 & 11.3 & 28.9 & 48.8 & 54.5 \\
\;\;Greedy based on fill rate & 18.3 & 31.6 & 26.1 & 15.0 & 19.6 & 18.8 & 13.4 & 7.4 & 11.8 & 14.3 & 15.5 \\
\;\;Alternate greedy storage and RL & 13.9 & 41.6 & 60.6 & 13.5 & 32.5 & 49.8 & 74.2 & 7.4 & 20.6 & 29.5 & 45.3 \\
\;\;Multiple sequence alignment & 30.4 & 49.2 & 41.5 & 17.9 & 19.9 & 5.9 & \textendash$^{\rm{b}}$ & 11.5 & 12.8 & 2.6 & \textendash$^{\rm{b}}$ \\
\;\;Simulated annealing & 18.3 & 39.5 & 50.2 & 14.0 & 23.0 & 19.4 & 13.2 & 8.1 & 12.1 & 15.0 & \B 1.2 \\

\bottomrule
\multicolumn{11}{l}{$^{\rm{a}}$ one out of ten instances not solved in cutoff time, $^{\rm{b}}$ no instances solved in cutoff time}\\

\end{tabular}
\end{table}


The results are shown in \Cref{tbl:filled_buffer}. We find that several approaches can no longer solve the larger instances, denoted by ``\textendash''.
We find that, on average, RL achieves the best solutions for most buffer sizes as it is only outperformed by simulated annealing for 15 colors and 8x8 buffer. Gurobi reaches the smallest ARPD among all baselines for 2x2 buffers, whereas Greedy based on fill rate and multiple sequence alignment are competitive for 6x6 and 8x8 buffers. \REV{The statistical tests again suggest significant performance differences, see Appendix K of the supplementary material.}

\subsection{Out-of-distribution instances}

We now assess whether the performance of our RL approach remains robust against out-of-distribution problem instances. Accordingly, we used the trained policies from the main analysis based on balanced color distributions ($p_1,\dots, p_C = 1/C$) and applied them to imbalanced color distributions. We consider imbalanced color distributions with exponentially or linearly decreasing probabilities. The exponentially decreasing probabilities decay by a factor of $0.8$. Hence, we have $p_1=\frac{1}{\sum_{i=0}^{C-1} 0.8^{i}}=\frac{0.2}{1-0.8^C}$ and $p_{i}=0.8\,p_{i-1}$, for $i=2,\dots,C$. 
For the linearly decreasing probabilities, we define $p_1=\frac{1}{C}+d(C)\frac{C-1}{2}$ and $p_i-p_{i-1}=d(C)$ for $i=2,\dots,C$ with slope $d(C)$ depending on the number of colors. We chose $d(5)= 0.08$, $d(10)=0.02$ and $d(15)=0.008$.
Note that the slope $d(C)$ needs to account for normalization and positivity so that $d(C) < \frac{2}{C(C-1)}$.\footnote{We need $\sum_{i=1}^C p_i = 1$ by definition of probabilities. Accordingly, we get $\sum_{i=1}^C p_i = \sum_{i=1}^C (p_1-(i-1)d(C))=Cp_1-d(C)\sum_{i=1}^C(i-1)=C p_1-d(C) \frac{C(C-1)}{2}$. Thus, we have $p_1=\frac{1}{C}+d(C)\frac{C-1}{2}$. All probabilities must be positive, in particular, we need $p_C>0 $ for the color with the smallest probability. Hence, we require $p_C = p_1-(C-1)d(C)=\frac{1}{C}+d(C)\frac{C-1}{2}-(C-1)d(C)=\frac{1}{C}-d(C)\frac{C-1}{2} > 0$, which implies $d(C)<\frac{2}{C(C-1)}$.}


\begin{table}[H]
\setlength{\tabcolsep}{2.5pt}
\caption{Evaluation results (ARPD) for imbalanced color distribution with exponentially decreasing probabilities.}
\label{tbl:exponential}
\footnotesize
\centering
\vspace{-0.2cm}
\begin{tabular}{l*{11}{S[detect-weight, mode=text,table-format=2.1,table-column-width=0.8cm]}}
\toprule
Colors & \multicolumn{3}{c}{5} & \multicolumn{4}{c}{10} & \multicolumn{4}{c}{15} \\
 \cmidrule(lr){2-4} \cmidrule(lr){5-8} \cmidrule(lr){9-12}
Buffer size (lanes x lane width) &    {2x2} &     {3x3} &     {4x4} &     {2x2} &   {4x4} &   {6x6} &   {8x8}  &    {2x2} &   {4x4} &   {6x6} &   {8x8} \\
\midrule
RL all action masks (sampling) & \B 1.1 & \B 0.0 & \B 0.0 & \B 1.5 & \B 0.2 & \B 1.5 & 14.3 & \B 3.2 & \B 2.1 & 9.9 & 23.7 \\

\addlinespace
\multicolumn{11}{l}{\underline{Baselines}} \\ \addlinespace

\;\;Gurobi (our ILP formulation) & 9.2 & 105.6 & 190.4 & 4.7 & 82.0 & 199.2 & 372.2 & 6.5 & 63.1 & 131.9 & 241.8 \\
\;\;Gurobi \citep{wu2021mathematical} & 22.7 & 35.8 & 26.4 & 14.5 & 8.9 & 102.5 & 251.0 & 14.5 & 8.0 & 80.1 & 177.2 \\
\;\;Greedy & 22.7 & 64.5 & 76.8 & 9.7 & 40.4 & 65.9 & 65.7 & 11.6 & 28.8 & 49.0 & 63.6 \\
\;\;Greedy based on fill rate & 11.9 & 28.3 & 30.0 & 4.3 & 10.3 & 12.8 & \B 1.7 & 5.7 & 9.1 & 10.7 & 19.9 \\
\;\;Alternate greedy storage and RL & 15.3 & 32.8 & 31.9 & 10.1 & 18.4 & 20.0 & 12.9 & 13.9 & 14.0 & 23.6 & 37.3 \\
\;\;Multiple sequence alignment & 31.4 & 58.6 & 61.1 & 17.9 & 23.2 & 15.3 & 17.9 & 17.0 & 22.7 & 20.4 & 27.5 \\
\;\;Simulated annealing & 18.9 & 41.6 & 47.5 & 4.4 & 10.3 & 16.1 & 16.9 & 8.2 & 5.2 & \B 0.3 & \B 0.0 \\
\bottomrule
\end{tabular}
\end{table}


The results are presented in \Cref{tbl:exponential} and \Cref{tbl:linear} for exponentially and linearly decreasing probabilities, respectively. We find that RL achieves the best solutions for most buffer sizes. However, RL is outperformed by Greedy based on fill rate for 10 colors and 8x8 buffers in both analyses. For exponentially decreasing probabilities, SA outperforms all approaches on problems with 15 colors and buffer sizes greater or equal to 5x5. For linearly decreasing probabilities, SA dominates on problems with 15 colors and buffer sizes greater or equal to 6x6. Similarly to our previous analyses, Gurobi presents the best baseline for 2x2 buffers. \REV{The statistical tests suggest a significant Friedman statistic in both analyses ($p < 0.001$). However, the Nemenyi test of \emph{RL vs Greedy based on fill rate} is not statistically significant with a $p$-value slightly above 5\% ($p=0.054$) for linearly decaying probabilities (see Appendix K).} The results on out-of-distribution instances point towards the robustness of RL as it remains superior in most problem settings with $\text{ARPD} \leq 33.8$.


\begin{table}[H]
\setlength{\tabcolsep}{3pt}
\caption{Evaluation results (ARPD) for imbalanced color distribution with linearly decreasing probabilities.}
\label{tbl:linear}

\footnotesize
\centering
\vspace{-0.2cm}
\begin{tabular}{l*{11}{S[detect-weight, mode=text,table-format=2.1,table-column-width=0.8cm]}}
\toprule
Colors & \multicolumn{3}{c}{5} & \multicolumn{4}{c}{10} & \multicolumn{4}{c}{15} \\
 \cmidrule(lr){2-4} \cmidrule(lr){5-8} \cmidrule(lr){9-12}
Buffer size (lanes x lane width) &    {2x2} &     {3x3} &     {4x4} &     {2x2} &   {4x4} &   {6x6} &   {8x8}  &    {2x2} &   {4x4} &   {6x6} &   {8x8} \\
\midrule
\addlinespace
RL all action masks (sampling)        & \B 0.5 & \B 0.0 & \B 0.0 & \B 3.8 & \B 0.5 & \B 1.6 & 12.9 & \B 5.5 & \B 1.7 & 13.0 & 33.8 \\

\addlinespace
\multicolumn{11}{l}{\underline{Baselines}} \\ \addlinespace

\;\;Gurobi (our ILP formulation) & 10.9 & 96.6 & 215.0 & 5.1 & 85.9 & 203.8 & 359.4 & 5.9 & 41.5 & 132.8 & 252.4 \\
\;\;Gurobi \citep{wu2021mathematical} & 33.9 & 43.5 & 32.8 & 19.8 & 9.2 & 119.9 & 264.1 & 12.5 & 3.5 & 71.0 & 169.9 \\
\;\;Greedy & 25.6 & 57.5 & 80.5 & 18.7 & 44.0 & 71.6 & 68.6 & 13.4 & 26.6 & 54.8 & 75.3 \\
\;\;Greedy based on fill rate & 17.8 & 29.6 & 16.6 & 12.4 & 13.0 & 10.4 & \B 3.5 & 10.0 & 7.3 & 16.6 & 29.2 \\
\;\;Alternate greedy storage and RL & 18.4 & 40.3 & 36.3 & 13.9 & 25.4 & 21.1 & 17.8 & 13.0 & 14.9 & 27.5 & 45.1 \\
\;\;Multiple sequence alignment & 42.0 & 66.6 & 68.0 & 23.5 & 29.8 & 20.6 & 13.1 & 15.2 & 16.8 & 18.0 & 34.6 \\
\;\;Simulated annealing & 21.5 & 46.1 & 51.6 & 12.8 & 13.3 & 16.3 & 16.5 & 10.5 & 7.2 & \B 2.2 & \B 0.0 \\

\bottomrule
\end{tabular}
\end{table}

\subsection{Longest sequences of exclusive storage or retrieval operations}

\begin{figure}[H]
\begin{subfigure}[b]{0.33\textwidth}
    \centering
    \begin{tikzpicture}
    \begin{axis}[
        width=\linewidth,
        height=6cm,
        xlabel={},
        boxplot/draw direction=y,
        ylabel={Maximum exclusive\\ sequence length},
        ylabel style={font=\footnotesize, text width=3cm, align=center},
        xmin=-0.5, xmax=3.5,
        ymin=0, ymax=67,
        xtick={0, 1, 2, 3},
        xticklabels={$\text{RL}^{\text{det}}$ , $\text{RL}^{\text{sto}}$, SA, Gurobi\\\citepalias{wu2021mathematical}},
        xticklabel style={font=\scriptsize, align=center},
        yticklabel style={font=\scriptsize},
        grid=major,
        grid style={dashed, gray!50},
        title style={yshift=0.5cm}
    ]
    
    \addplot[only marks, mark=*, mark size=2pt, blue] coordinates {
        (0, 10)
        (0, 11)
        (0, 8)
        (0, 14)
        (0, 9)
        (0, 12)
        (0, 13)
        (0, 12)
        (0, 15)
        (0, 17)
    };

 \addplot[
    color=black,              
    draw=black,             
    fill=gray!30,             
    boxplot prepared={
        lower whisker=5,
        upper whisker=26,
        lower quartile=9,
        upper quartile=13,
        median=11,
        draw position=1
    }
] coordinates {};

	\addplot[only marks, mark=*, mark size=2pt, green] coordinates {
        (2, 15)
        (2, 10)
        (2, 26)
        (2, 13)
        (2, 17)
        (2, 15)
        (2, 15)
        (2, 12)
        (2, 12)
        (2, 11)
                
    };      
    
    \addplot[only marks, red, mark size=2pt] coordinates {(3, 36)};
    
    \end{axis}
    
    \end{tikzpicture}
    \caption{\footnotesize 6x6 buffer.}
    \end{subfigure}
\begin{subfigure}[b]{0.33\textwidth}
    \centering
    \begin{tikzpicture}
    \begin{axis}[
    boxplot/draw direction=y,
         width=\linewidth,
        height=6cm,
        xlabel={},
        ylabel={},
        ylabel style={font=\footnotesize, draw=none},
        xmin=-0.5, xmax=3.5,
        ymin=0, ymax=67,
        xtick={0, 1, 2, 3},
        xticklabels={$\text{RL}^{\text{det}}$ , $\text{RL}^{\text{sto}}$, SA, Gurobi\\\citepalias{wu2021mathematical}},
        xticklabel style={font=\scriptsize, align=center},
        yticklabel style={font=\scriptsize},
        grid=major,
        grid style={dashed, gray!50},
        title style={yshift=0.5cm}
    ]

    \addplot[only marks, mark=*, mark size=2pt, blue] coordinates {
        (0, 15)
        (0, 10)
        (0, 26)
        (0, 13)
        (0, 17)
        (0, 15)
        (0, 15)
        (0, 12)
        (0, 12)
        (0, 11)
    };

    \addplot[
    color=black,              
    draw=black,               
    fill=gray!30,             
    boxplot prepared={
        lower whisker=6,
        upper whisker=35,
        lower quartile=12,
        upper quartile=17,
        median=13,
        draw position=1
    }
] coordinates {}; 
    
	\addplot[only marks, mark=*, mark size=2pt, green] coordinates {
        (2, 18)
        (2, 26)
        (2, 32)
        (2, 24)
        (2, 17)
        (2, 21)
        (2, 21)
        (2, 23)
        (2, 18)
        (2, 23)

    };    
    
    \addplot[only marks, red, mark size=2pt] coordinates {(3, 49)};
    
    \end{axis}
    
    \end{tikzpicture}
    \caption{\footnotesize 7x7 buffer.}
    \end{subfigure}
    \begin{subfigure}[b]{0.33\textwidth}
    \centering
    \begin{tikzpicture}
    \begin{axis}[
    boxplot/draw direction=y,
   		  width=\linewidth,
        height=6cm,
        xlabel={},
        ylabel={},
        ylabel style={font=\footnotesize, draw=none},
        xmin=-0.5, xmax=3.5,
        ymin=0, ymax=67,
        xtick={0, 1, 2, 3},
        xticklabels={$\text{RL}^{\text{det}}$ , $\text{RL}^{\text{sto}}$, SA, Gurobi\\\citepalias{wu2021mathematical}},
        xticklabel style={font=\scriptsize, align=center},
        yticklabel style={font=\scriptsize},
        grid=major,
        grid style={dashed, gray!50},
        title style={yshift=0.5cm}
    ]
    
    \addplot[only marks, mark=*, mark size=2pt, blue] coordinates {
        (0, 13)
        (0, 32)
        (0, 14)
        (0, 18)
        (0, 12)
        (0, 18)
        (0, 14)
        (0, 9)
        (0, 16)
        (0, 17)
    };

        \addplot[
    color=black,             
    draw=black,               
    fill=gray!30,            
    boxplot prepared={
        lower whisker=6,
        upper whisker=41,
        lower quartile=12,
        upper quartile=17,
        median=14,
        draw position=1
    }
] coordinates {}; 
    
	\addplot[only marks, mark=*, mark size=2pt, green] coordinates {
        (2, 27)
        (2, 22)
        (2, 39)
        (2, 20)
        (2, 16)
        (2, 18)
        (2, 24)
        (2, 27)
        (2, 37)
        (2, 20)

    };      
    
    \addplot[only marks, red, mark size=2pt] coordinates {(3, 64)};
    
    \end{axis}
    
    \end{tikzpicture}
    \caption{\footnotesize 8x8 buffer.}
    \end{subfigure}    
    
    \caption{Comparison of the longest exclusive sequences of storage or retrieval operations between RL, simulated annealing, and Gurobi based on ``store-then-retrieve'' \cite{wu2021mathematical}. Each plot shows the maximum exclusive sequence lengths of $n=10$ instances with 15 colors. The first and last sub-sequence was excluded in the calculation of the maximum to remove initial filling and final retrieval.}
    \label{fig:exclusive}
\end{figure}
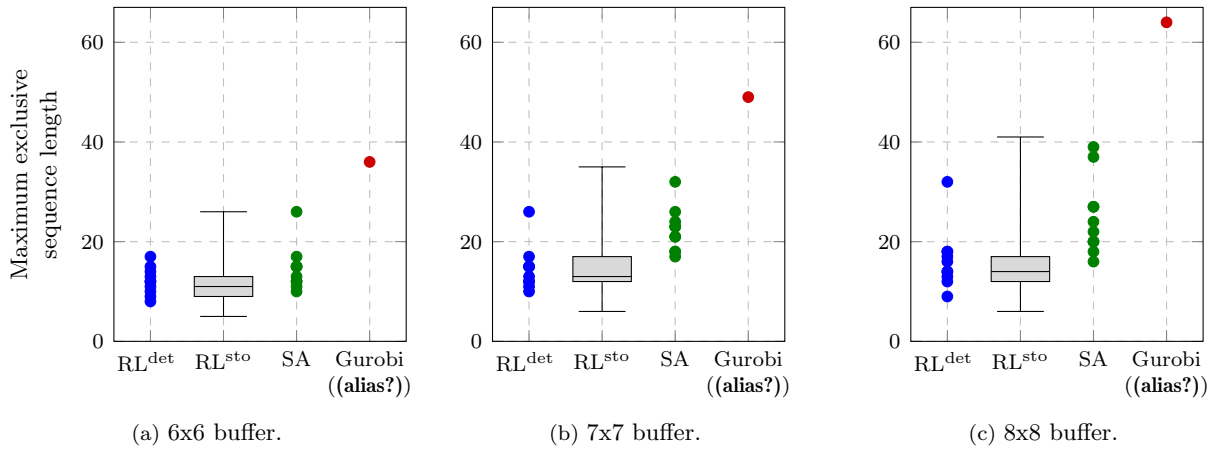

We finally analyze the prevalence of long sequences of exclusive storage or retrieval operations. Such long exclusive sequences can interrupt a constant flow of workpieces, leading to undesirable effects (e.g., additional space requirements, waiting times). Therefore, the practicability of the solutions might not only depend on the number of color changes but also on the lengths of exclusive sequences. \Cref{fig:exclusive} shows the maximum length of subsequences of exclusive storage or retrieval operations in the solutions of several approaches for different buffer sizes. We only considered the problem instances with initially filled buffer. We included RL with deterministic policy application and repeated sampling. The results for RL with repeated sampling are based on all sequences generated in 5 minutes (approximately a thousand) for each instance and hence summarized as a boxplot, with maximum and minimum as upper and lower whiskers. We find that the maximum lengths of exclusive storage or retrieval sequences generated by the RL approaches with deterministic policy application and repeated sampling remain, on average, considerably below those produced by simulated annealing and the Gurobi solver based on the ``store-then-retrieve'' variant \citep{wu2021mathematical}. Accordingly, the solutions generated by RL achieve a low number of color changes, while also ensuring a constant flow of workpieces through the multi-lane buffer.

\section{Conclusion}
\label{section-Conclusion}


We proposed a reinforcement learning approach for the paint shop problem with multi-lane buffers that allows fully flexibility in performing store and retrieve operations. To the best of our knowledge, our study is the first to provide an ILP formalization of this problem variant. We showed formally that the less flexible problem variant ``store-then-variant'' can lead to solutions that are arbitrarily worse than our flexible problem variant allowing store and retrieve operations to be performed in an arbitrary order. We then present a RL approach based on proximal policy optimization that is trained to minimize color changes. In contrast to prior works that proposed RL approaches for the paint shop problem with multi-lane buffers \citep{Huang.2024,Leng.2020,Leng.2023}, we employ RL to perform both, store and retrieve operations. 
The one-hot encoded state representation describes the buffer content, the colors of the most recent cars from the upstream sequence, and the color of the next car in the downstream sequence. After showing that greedy retrieval and fast-track actions are optimal, we employed action masking to incorporate this prior knowledge into policy learning and application. Our evaluation based on 170 problem instances suggests that the proposed RL approach reduces color changes for most considered problem instances by considerable margins. In addition, we demonstrated the robustness of our approach towards rectangular buffer sizes, initially filled buffers, and out-of-distribution instances.


Our evaluation suggests several guidelines for implementing reinforcement learning approaches to the paint shop problem with multi-lane buffers. First, researchers should include action masking in policy learning as it leads to a more efficient learning process by directly enforcing valid and provably optimal actions. We presented several action masks and showed how they can be combined into a single action mask with a defined priority structure. 
Second, we found that repeated sampling from the trained policy yields significantly better solution quality than applying the policy deterministically. Since a trained policy can generate solutions very quickly, the limited cutoff time can be used effectively to sample and evaluate multiple solutions.
Third, our results showed that a policy trained on problem instances with balanced color distributions also performs well on out-of-distribution instances with imbalanced color distributions, suggesting robustness towards unforeseen changes in problem parameters. Finally, we found that deterministic and stochastic RL application also results in considerably shorter  sequences of exclusive storage or retrieval operations than simulated annealing and methods based on ``store-then-retrieve'', thereby ensuring a more continuous and stable flow of workpieces in production.


Our study also provides several opportunities for future research. First, one could consider more complex problem variants, e.g., with more than one painting line. If multiple painting lines are employed, the downstream sequence can be split further into multiple downstream sequences, which reduces the number of color changes. Second, other performance metrics, like makespan of individual cars could be considered in addition to color changes. So far, we have allowed each car to remain in the buffer for an arbitrary amount of time without considering any constraints regarding makespan. The model would then have to learn to account for constrained makespans by performing timely retrieve actions. Third, future research could attempt to integrate the paint shop problem into other scheduling problems like mixed-model sequencing or flow shop problems. Here, researchers first need to weigh the performance metrics of the main scheduling problem against the number of color changes. Another challenge is to quickly estimate the number of color changes of a sequence under a (near-)optimal action sequence in the paint shop. 

\singlespacing
\footnotesize
\bibliography{references} 

@article{lahmar2003resequencing,
  title={Resequencing and feature assignment on an automated assembly line},
  author={Lahmar, Maher and Ergan, Hakan and Benjaafar, Saif},
  journal={IEEE Transactions on Robotics and Automation},
  volume={19},
  number={1},
  pages={89--102},
  year={2003},
  publisher={IEEE}
}

@article{stable-baselines3,
  author  = {Antonin Raffin and Ashley Hill and Adam Gleave and Anssi Kanervisto and Maximilian Ernestus and Noah Dormann},
  title   = {Stable-Baselines3: {R}eliable Reinforcement Learning Implementations},
  journal = {Journal of Machine Learning Research},
  year    = {2021},
  volume  = {22},
  number  = {268},
  pages   = {1-8},
  url     = {http://jmlr.org/papers/v22/20-1364.html}
}

@article{Friedman.1937,
  title={The use of ranks to avoid the assumption of normality implicit in the analysis of variance},
  author={Friedman, Milton},
  journal={Journal of the American Statistical Association},
  volume={32},
  number={200},
  pages={675--701},
  year={1937},
  publisher={Taylor \& Francis}
}

@phdthesis{Nemenyi.1963,
  title={Distribution-free multiple comparisons},
  author={Nemenyi, Peter Bjorn},
  year={1963},
  school={Princeton University}
}

@article{Bysko.2024,
  title={Nash equilibrium as a tool for the Car Sequencing Problem 4.0},
  author={Bysko, Sara and Krystek, Jolanta and {\'S}wierniak, Andrzej},
  journal={Journal of Intelligent Manufacturing},
  volume={35},
  number={3},
  pages={1037--1053},
  year={2024},
  publisher={Springer}
}

@article{Panzer.2022,
  title={Deep reinforcement learning in production systems: {A} systematic literature review},
  author={Panzer, Marcel and Bender, Benedict},
  journal={International Journal of Production Research},
  volume={60},
  number={13},
  pages={4316--4341},
  year={2022},
  publisher={Taylor \& Francis}
}

@article{Gunay.2017,
  title={A stochastic programming model for resequencing buffer content optimisation in mixed-model assembly lines},
  author={Gunay, Elif Elcin and Kula, Ufuk},
  journal={International Journal of Production Research},
  volume={55},
  number={10},
  pages={2897--2912},
  year={2017},
  publisher={Taylor \& Francis}
}

@article{Huang.2024,
  title={Deep reinforcement learning for solving car resequencing with selectivity banks in automotive assembly shops},
  author={Huang, Yuzhe and Fu, Gaocai and Sheng, Buyun and Lu, Yingkang and Yu, Junpeng and Yin, Xiyan},
  journal={International Journal of Production Research},
  pages={2363--2384},
  volume={63},
  number={7},
  year={2025},
  publisher={Taylor \& Francis}
}

@article{Boysen.2011,
  title={The car resequencing problem with pull-off tables},
  author={Boysen, Nils and Golle, Uli and Rothlauf, Franz},
  journal={Business Research},
  volume={4},
  number={2},
  pages={276--292},
  year={2011},
  publisher={Springer}
}

@article{Valero.2014,
  title={Solving the Car Resequencing Problem with mix banks},
  author={Valero-Herrero, Maria and Garc{\'\i}a Sabater, Jos{\'e} Pedro and Vidal-Carreras, Pilar I and Can{\'o}s-Dar{\'o}s, Lourdes},
  journal={Direcci{\'o}n y Organizaci{\'o}n},
  volume={54},
  pages={36--44},
  year={2014},
  publisher={Universidad Polit{\'e}cnica de Madrid}
}

@article{Boysen.2013,
title = {A decomposition approach for the car resequencing problem with selectivity banks},
journal = {Computers \& Operations Research},
volume = {40},
number = {1},
pages = {98--108},
year = {2013},
author = {Nils Boysen and Michael Zenker},
}

@article{Sun.2024,
title = {Integrating virtual resequencing with car resequencing via selectivity banks for mixed-model assembly lines},
journal = {Computers \& Industrial Engineering},
volume = {189},
pages = {109990},
year = {2024},
author = {Hui Sun},
}

@article{inman2003algorithm,
  title={Algorithm for agile assembling-to-order in the automotive industry},
  author={Inman, Robert R and Schmeling, DM},
  journal={International Journal of Production Research},
  volume={41},
  number={16},
  pages={3831--3848},
  year={2003},
  publisher={Taylor \& Francis}
}

@article{taube2018resequencing,
  title={Resequencing mixed-model assembly lines with restoration to customer orders},
  author={Taube, F and Minner, Stefan},
  journal={Omega},
  volume={78},
  pages={99--111},
  year={2018},
  publisher={Elsevier}
}

@article{goodson2015priori,
  title={A priori policy evaluation and cyclic-order-based simulated annealing for the multi-compartment vehicle routing problem with stochastic demands},
  author={Goodson, Justin C},
  journal={European Journal of Operational Research},
  volume={241},
  number={2},
  pages={361--369},
  year={2015},
  publisher={Elsevier}
}

@article{knopp2017batch,
title = {A batch-oblivious approach for Complex Job-Shop scheduling problems},
journal = {European Journal of Operational Research},
volume = {263},
number = {1},
pages = {50-61},
year = {2017},
issn = {0377-2217},
doi = {https://doi.org/10.1016/j.ejor.2017.04.050},
url = {https://www.sciencedirect.com/science/article/pii/S0377221717303971},
author = {Sebastian Knopp and Stéphane Dauzère-Pérès and Claude Yugma},
}

@book{gendreau2018handbook,
  title={Handbook of Metaheuristics},
  author={Gendreau, Michel and Potvin, Jean-Yves},
  volume={272},
  year={2018},
  publisher={Springer},
  address = {New York}
}

@misc{psproblemdata,
  title={Data for: Solving the Paint Shop Problem Using Reinforcement Learning},
  author={Stappert, Mirko and Lutz, Bernhard and Brammer, Janis and Neumann, Dirk},
  howpublished = "Dataset on Mendeley",
  doi = {10.17632/zbg64f6vb3.1},
  note = {Available at \url{https://doi.org/10.17632/zbg64f6vb3.1}},
  year={2023}
}

@article{wu2021mathematical,
  title={Mathematical modeling and heuristic approaches for a multi-stage car sequencing problem},
  author={Wu, Jiaxi and Ding, Yongkang and Shi, Leyuan},
  journal={Computers \& Industrial Engineering},
  volume={152},
  pages={107008},
  year={2021},
  publisher={Elsevier}
}

@article{sun2015colour,
  title={A colour-batching problem using selectivity banks in automobile paint shops},
  author={Sun, Hui and Fan, Shujin and Shao, Xianle and Zhou, Jiangong},
  journal={International Journal of Production Research},
  volume={53},
  number={4},
  pages={1124--1142},
  year={2015},
  publisher={Taylor \& Francis}
}

@article{Plastria2002formulating,
title = {Formulating logical implications in combinatorial optimisation},
journal = {European Journal of Operational Research},
volume = {140},
number = {2},
pages = {338-353},
year = {2002},
issn = {0377-2217},
doi = {https://doi.org/10.1016/S0377-2217(02)00073-5},
url = {https://www.sciencedirect.com/science/article/pii/S0377221702000735},
author = {Frank Plastria}
}

@article{ko2016paint,
  title={Paint batching problem on {M}-to-1 conveyor systems},
  author={Ko, Sung-Seok and Han, Yong-Hee and Choi, Jin Young},
  journal={Computers \& Operations Research},
  volume={74},
  pages={118--126},
  year={2016},
  publisher={Elsevier}
}

@article{sun2017study,
  title={A study on implementing color-batching with selectivity banks in automotive paint shops},
  author={Sun, Hui and Han, Jianming},
  journal={Journal of Manufacturing Systems},
  volume={44},
  pages={42--52},
  year={2017},
  publisher={Elsevier}
}

@article{Mosadegh.2020,
	title = {Stochastic mixed-model assembly line sequencing problem: {Mathematical} modeling and {Q}-learning based simulated annealing hyper-heuristics},
	journal = {European Journal of Operational Research},
	author = {Mosadegh, H and Fatemi Ghomi, SMT and S\"{u}er, GA},
	year = {2020},
	volume = {282},
	number = {2},
	pages = {530-544}
}

@article{Leng.2020,
  title={Deep reinforcement learning for a color-batching resequencing problem},
  author={Leng, Jinling and Jin, Chun and Vogl, Alexander and Liu, Huiyu},
  journal={Journal of Manufacturing Systems},
  volume={56},
  pages={175--187},
  year={2020},
  publisher={Elsevier}
}

@article{Leng.2023,
  title={A multi-objective reinforcement learning approach for resequencing scheduling problems in automotive manufacturing systems},
  author={Leng, Jinling and Wang, Xingyuan and Wu, Shiping and Jin, Chun and Tang, Meng and Liu, Rui and Vogl, Alexander and Liu, Huiyu},
  journal={International Journal of Production Research},
  volume={61},
  number={15},
  pages={5156--5175},
  year={2023},
  publisher={Taylor \& Francis}
}

@incollection{Epping.2003,
  title={Sorting with line storage systems},
  author={Epping, Thomas and Hochst{\"a}ttler, Winfried},
  booktitle={Operations Research Proceedings 2002},
  pages={235--240},
  year={2003},
  publisher={Springer},
  address = {Berlin},
}

@inproceedings{Lin.2011,
  title={A research of resequencing problem in automobile paint shops using selectivity banks},
  author={Lin, Long and Sun, Hui and Xu, Ying-qiu},
  booktitle={International Conference on Industrial Engineering and Engineering Management},
  pages={658--662},
  year={2011},
  organization={IEEE}
}

@article{Ding.2004,
  title={Sequence alteration and restoration related to sequenced parts delivery on an automobile mixed-model assembly line with multiple departments},
  author={Ding, F-Y and Sun, Hui},
  journal={International Journal of Production Research},
  volume={42},
  number={8},
  pages={1525--1543},
  year={2004},
  publisher={Taylor \& Francis}
}

@article{Neufeld.2023,
  title={A systematic review of multi-objective hybrid flow shop scheduling},
  author={Neufeld, Janis S and Schulz, Sven and Buscher, Udo},
  journal={European Journal of Operational Research},
  volume={309},
  number={1},
  pages={1--23},
  year={2023},
  publisher={Elsevier}
}

@article{Brammer.2022a,
  title={Permutation flow shop scheduling with multiple lines and demand plans using reinforcement learning},
  author={Brammer, Janis and Lutz, Bernhard and Neumann, Dirk},
  journal={European Journal of Operational Research},
  volume={299},
  number={1},
  pages={75--86},
  year={2022},
  publisher={Elsevier}
}

@article{Brammer.2021,
  title={Solving the mixed model sequencing problem with reinforcement learning and metaheuristics},
  author={Brammer, Janis and Lutz, Bernhard and Neumann, Dirk},
  journal={Computers \& Industrial Engineering},
  volume={162},
  pages={107704},
  year={2021},
  publisher={Elsevier}
}

@article{Brammer.2022b,
  title={Stochastic mixed model sequencing with multiple stations using reinforcement learning and probability quantiles},
  author={Brammer, Janis and Lutz, Bernhard and Neumann, Dirk},
  journal={OR Spectrum},
  volume={44},
  number={1},
  pages={29--56},
  year={2022},
  publisher={Springer}
}

@article{Boysen.2022,
  title={Assembly line balancing: {W}hat happened in the last fifteen years?},
  author={Boysen, Nils and Schulze, Philipp and Scholl, Armin},
  journal={European Journal of Operational Research},
  volume={301},
  number={3},
  pages={797--814},
  year={2022},
  publisher={Elsevier}
}

@article{Bysko.2020,
title = {Automotive Paint Shop 4.0},
journal = {Computers \& Industrial Engineering},
volume = {139},
pages = {105546},
year = {2020},
author = {Sara Bysko and Jolanta Krystek and Szymon Bysko}
}

@article{Bengio.2021,
title = {Machine learning for combinatorial optimization: {A} methodological tour d'horizon},
journal = {European Journal of Operational Research},
volume = {290},
number = {2},
pages = {405-421},
year = {2021},
author = {Yoshua Bengio and Andrea Lodi and Antoine Prouvost}
}

@article{boysen_resequencing_2012,
	title = {Resequencing of mixed-model assembly lines: {Survey} and research agenda},
	volume = {216},
	issn = {03772217},
	shorttitle = {Resequencing of mixed-model assembly lines},
	url = {http://linkinghub.elsevier.com/retrieve/pii/S0377221711007284},
	doi = {10.1016/j.ejor.2011.08.009},
	language = {en},
	number = {3},
	urldate = {2018-01-05},
	journal = {European Journal of Operational Research},
	author = {Boysen, Nils and Scholl, Armin and Wopperer, Nico},
	month = feb,
	year = {2012},
	pages = {594--604}
}

@article{spieckermann_sequential_2004,
	title = {A sequential ordering problem in automotive paint shops},
	volume = {42},
	issn = {0020-7543, 1366-588X},
	url = {https://www.tandfonline.com/doi/full/10.1080/00207540310001646821},
	doi = {10.1080/00207540310001646821},
	language = {en},
	number = {9},
	urldate = {2019-02-07},
	journal = {International Journal of Production Research},
	author = {Spieckermann, S. and Gutenschwager, K. and Voß, S.},
	month = may,
	year = {2004},
	pages = {1865--1878}
}

@misc{schulman_high-dimensional_2015,
	title = {High-dimensional continuous control using generalized advantage estimation},
	author = {Schulman, John and Moritz, Philipp and Levine, Sergey and Jordan, Michael and Abbeel, Pieter},
	year = {2015},
    note = {arXiv preprint, \url{https://arxiv.org/abs/1506.02438}}
}

@misc{schulman_proximal_2017,
	title = {Proximal policy optimization algorithms},
	author = {Schulman, John and Wolski, Filip and Dhariwal, Prafulla and Radford, Alec and Klimov, Oleg},
	year = {2017},
    note = {arXiv preprint, \url{https://arxiv.org/abs/1707.06347}}
}

@unpublished{brockman_openai_2016,
	title = {{OpenAI} {Gym}},
	author = {Brockman, Greg and Cheung, Vicki and Pettersson, Ludwig and Schneider, Jonas and Schulman, John and Tang, Jie and Zaremba, Wojciech},
	year = {2016},
    note = {Available at \url{https://arxiv.org/abs/1606.01540}}
}

@misc{gurobi,
  author = {{Gurobi Optimization, LLC}},
  title = {{Gurobi Optimizer Reference Manual}},
  year = 2023,
  note = {Available at \url{https://www.gurobi.com}}
}

@article{hong2018accelerated,
  title={Accelerated dynamic programming algorithms for a car resequencing problem in automotive paint shops},
  author={Hong, Sungwon and Han, Jinil and Choi, Jin Young and Lee, Kyungsik},
  journal={Applied Mathematical Modelling},
  volume={64},
  pages={285--297},
  year={2018},
  publisher={Elsevier}
}

\end{document}